\documentclass[a4paper,twoside,fleqn,10pt]{my-cas-dc}

\usepackage{hyperref} 

\usepackage[numbers]{natbib}

\usepackage{float}

\usepackage{amsfonts}
\usepackage{amsmath}
\usepackage{amssymb}

\usepackage{algorithmic}

\usepackage{array}

\usepackage[caption=false,font=footnotesize]{subfig}

\usepackage{environ}
\usepackage{pgfplots}
\usepackage{sansmath}
\usepackage{tikz}
\usepackage{tikz-3dplot}
\usepackage{upgreek}
\usepackage{url}
\usepackage[capitalize]{cleveref}

\usepackage{enumitem}

\pgfplotsset{compat=1.14}

\usepackage[absolute,overlay]{textpos}

\begin{document}

\begin{textblock*}{17.6cm}(1.6cm,0.5cm) 
   \noindent \footnotesize{This is an accepted manuscript preprint of published work available at \url{https://doi.org/10.1016/j.robot.2019.103380}. Please cite with Kirk Y.W Scheper and Guido C.H.E. de Croon, Evolution of robust high speed optical-flow-based landing for autonomous MAVs, Robotics and Autonomous Systems 124, 2020.}
\end{textblock*}

\let\WriteBookmarks\relax
\def\floatpagepagefraction{1}
\def\textpagefraction{.001}
\shorttitle{Robotics and Autonomous Systems}
\shortauthors{Scheper and de Croon}

\title [mode = title]{Evolution of Robust High Speed Optical-Flow-Based Landing for Autonomous MAVs}                      
\author{Kirk Y.W. Scheper}[orcid=0000-0003-2770-5556]
\cormark[1]
\cortext[cor1]{Corresponding author}
\ead{k.y.w.scheper@tudelft.nl}

\author{Guido C.H.E. de~Croon}[orcid=0000-0001-8265-1496]
\ead{g.c.h.e.Croon@tudelft.nl}



\address{Micro Air Vehicle Laboratory, Aerospace Engineering, Delft University of Technology, Delft, Netherlands}





\begin{abstract}[S U M M A R Y]
Automatic optimization of robotic behavior has been the long-standing goal of Evolutionary Robotics. Allowing the problem at hand to be solved by automation often leads to novel approaches and new insights. A common problem encountered with this approach is that when this optimization occurs in a simulated environment, the optimized policies are subject to the reality gap when implemented in the real world. This often results in sub-optimal behavior, if it works at all. This paper investigates the automatic optimization of neurocontrollers to perform quick but safe landing maneuvers for a quadrotor micro air vehicle using the divergence of the optical flow field of a downward looking camera. The optimized policies showed that a piece-wise linear control scheme is more effective than the simple linear scheme commonly used, something not yet considered by human designers. Additionally, we show the utility in using abstraction on the input and output of the controller as a tool to improve the robustness of the optimized policies to the reality gap by testing our policies optimized in simulation on real world vehicles. We tested the neurocontrollers using two different methods to generate and process the visual input, one using a conventional CMOS camera and one a dynamic vision sensor, both of which perform significantly differently than the simulated sensor. The use of the abstracted input resulted in near seamless transfer to the real world with the controllers showing high robustness to a clear reality gap.
\end{abstract}

\begin{keywords}
{
Initially Received 31 March 2019\\
Revised 15 October 2019\\
Accepted 15 November 2019}
{
Evolutionary Robotics \sep Bio-Inspired Landing \sep Reality Gap \sep High Speed Flight
}
\end{keywords}

\begin{highlights}
\begin{itemize}[leftmargin=*]
    \item Neurocontrollers for high speed optical-flow-based landing were automatically developed with better results than the user-defined baseline.
    \item The optimized behavior was seamlessly transferred to the real world with the aid of a closed loop control system.
    \item Preprocessing the sensory input allowed the behavior to be tested with both a CMOS camera and Dynamic Vision Sensor with no noticeable changes to the performance.
    \item No significant performance differences were observed despite clear differences in the neurocontroller input.
\end{itemize}
\end{highlights}

\maketitle

%
%
\section{Introduction}
Insects are a significant source of inspiration for developing novel solutions to autonomous flight of Micro Air Vehicles (MAVs). Even with limited computational and energy resources, insects are able to effectively complete challenging tasks with relatively complex behaviors. One behavior that is of particular interest, is landing.

Insects have been shown to primarily rely on visual inputs when landing, whereby many insects regulate a constant rate of expansion (or \emph{divergence}) of optical flow \cite{Gibson1950} to perform a smooth landing \cite{Baird2013}. This approach has inspired some robotic implementations of constant divergence landing strategies \cite{Herisse2012}. Although simple in concept, this approach is quite difficult to implement in reality. A direct implementation causes the robot to become unstable as the vehicle nears the ground. This is a result of the non-linear interaction between the vehicle control and sensing, in the presence of delay, measurement noise and environmental disturbance \cite{Croon2016}. Different augmentations to the standard control scheme have been made, most, simply perform slow landings \cite{Herisse2012}, use long landing legs to delay the onset of this instability and touch the ground before they occur or by switching to alternative measures like time-to-contact \cite{Clady2014,Kendoul2014} or velocity in-plane of the camera \cite{Rodolfo2015,Miller2019}. More recently, there have been attempts to identify the instability and adapt the landing strategy by adjusting control gains \cite{PijnackerHordijk2018,Ho2018}.

An alternative to these manually designed approaches is to have an optimization technique automatically develop a suitable solution that is robust to these instabilities. This optimization may reveal new solutions to this problem, and perhaps even alternative hypotheses on what flying insects such as honeybees may be doing. Some attempts have been made to do this \cite{Howard2016} but to the best knowledge of the authors, none of these have been implemented on real world robots.

This is due in part, to the effects of the differences between the simulated environment, commonly used for behavioral optimization, and the real world. This resultant difference in the robotic behavior is commonly referred to as the \emph{reality gap} \cite{jakobi1995noise, Nolfi2000, Bongard2013}. Several approaches have been used to make controllers more robust to the reality gap with the most significant being adding appropriate and varied noise \cite{Jakobi1995}, co-evaluation of controllers in the real world to test transferability \cite{Koos2010} and inspired by conventional control theory, there are approaches utilizing abstracted outputs from the neurocontroller with a closed loop control system to actively reduce the effect of reality gap \cite{Szczawinski2013, Scheper2016a}.

In this paper, we describe a method to optimize a quick but safe landing maneuver for a quadrotor MAV equipped with a downward facing camera. The neurocontrollers are given only the divergence of the optical flow field from the camera as input and its time derivative. Although this abstracted input may reduce the possible behaviors the controllers can express, building on the work in \cite{Scheper2017}, we will demonstrate that this abstraction leads to a robust transfer from simulation to reality after virtual optimization.

The following consists of a summary definition of optical flow in \cref{sec:opticalflow}. The flight platform and simulation environment is then described in \cref{sec:platform}. Next, the performance of conventional constant divergence landing approaches are presented in \cref{sec:baseline} to provide some baseline performance to compare the optimized policies against. The evolutionary setup and neural models used for the neurocontrollers is then described in \cref{sec:evo}. This is followed by a presentation and analysis of the optimized policies in \cref{sec:evo_results}. The reality gap and the results from the real world experiments are presented in \cref{sec:realityGap} and \cref{sec:flightresults} from which we draw some conclusions in \cref{sec:conclusion}.

%
%
\section{Optical Flow Definition}
\label{sec:opticalflow}

To perform optical flow based landing as in \cite{Croon2013a, Ho2016, PijnackerHordijk2018, Ho2018}, we must first define the optical flow parameters. The formulation here is a summarized version of that presented in \cite{Croon2013a}. This algorithm provides a good trade-off of accurate optical flow estimates while using relatively limited computational resources. This allows the perception and control loop to operate at high frequency and low latency on the embedded flight platform used in this paper. Alternative optical flow estimation methods could be used given that the estimation runs fast enough to facilitate the flight control. An investigation of the accuracy and reliability of the optical flow estimation is out of the scope of this paper.

\begin{figure}[pos=!tb]
	\centering
	\includegraphics[width=0.4\textwidth]{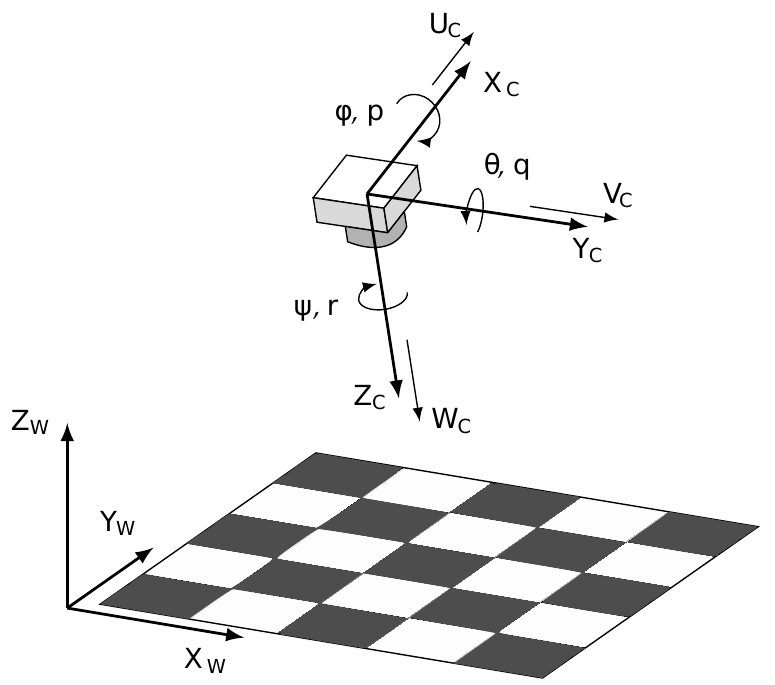}
	\caption{Definitions of the world ($\mathcal{W}$) and camera ($\mathcal{C}$) references frames. Also shown are the Euler angles ($\phi$, $\theta$, $\psi$), rotational rates ($p$, $q$ $r$), and translational velocities ($U$, $V$, $W$) that describe the motion of $\mathcal{C}$ \cite{ParedesValles2018}.}
	\label{fig:opticalflow_model}
\end{figure}

If we assume that we have a downward-looking camera overlooking a static planar scene, as shown in \cref{fig:opticalflow_model}, we can derive the perceived optical flow as the result of the camera ego-motion. The derivation of this optical flow model relies on the two reference frames, the inertial world frame is denoted by $\mathcal{W}$ and the camera frame centered at the focal point of the camera denoted by $\mathcal{C}$. In each of these frames, position is defined through the coordinates ($X$, $Y$, $Z$), with ($U$, $V$, $W$) as the corresponding velocity components. The orientation of $\mathcal{C}$ with respect to $\mathcal{W}$ is described by the Euler angles $\phi$, $\theta$, and $\psi$, denoting roll, pitch, and yaw, respectively. Similarly, $p$, $q$, and $r$ denote the corresponding rotational rates.

The camera ego motion can be related to the optical flow, and visual observables based on the pinhole camera model \cite{Longuet-Higgins1980} with camera pixel coordinates are denoted by ($x$, $y$), while ($u$,$v$) represent optical flow components, measured in pixels per second. These can be non-dimensionalized using the intrinsic calibration of the camera.
\begin{equation}\label{eqobs:1}
\begin{aligned}
u &= -\frac{U_{\mathcal{C}}}{Z_{\mathcal{C}}} + \frac{W_{\mathcal{C}}}{Z_{\mathcal{C}}}x - q + r y + p x y- q x^2\\
v &= -\frac{V_{\mathcal{C}}}{Z_{\mathcal{C}}} + \frac{W_{\mathcal{C}}}{Z_{\mathcal{C}}}y + p - r x - q x y + p y^2
\end{aligned}
\end{equation}
\cref{eqobs:1} shows that the optical flow of a point can be resolved into translational and rotational components. Since the latter is independent of the three-dimensional structure of the visual scene, these expressions can be derotated if information on the rotational rates of the camera is available. This derotation leads to pure translational optical flow components, denoted by ($u_{T}$, $v_{T}$). Moreover, if the scene is a planar surface, the depth $Z_{\mathcal{C}}$ of all visible world points are interrelated through: 
\begin{equation}\label{eqobs:2}
\begin{aligned}
Z_{\mathcal{C}} = Z_{0} + Z_{X}X_{\mathcal{C}} + Z_{Y}Y_{\mathcal{C}}
\end{aligned}
\end{equation}
\noindent{where $Z_{0}$ is defined as the distance to the surface along the optical axis of the camera, and $Z_{X}$ and $Z_{Y}$ represent the slopes of the planar scene with respect to the $X$- and $Y$-axis of $\mathcal{C}$.}

In \cite{Longuet-Higgins1980}, the relation between the position of an arbitrary point in $\mathcal{C}$ and its projection onto the image plane is given by ($x$, $y$) = ($X_{\mathcal{C}}/Z_{\mathcal{C}}$, $Y_{\mathcal{C}}/Z_{\mathcal{C}}$). Consequently, \cref{eqobs:2} may also be written in the form:
\begin{equation}\label{eqobs:4}
\begin{aligned}
\frac{Z_{\mathcal{C}}-Z_{0}}{Z_{\mathcal{C}}} = Z_{X} x + Z_{Y} y
\end{aligned}
\end{equation}
Further, let the \textit{scaled velocities} of the camera $\vartheta_{x}$, $\vartheta_{y}$, and $\vartheta_{z}$ be defined as follows:
\begin{equation}\label{eqobs:5}
\begin{aligned}
\vartheta_{x} = \frac{U_{\mathcal{C}}}{Z_{0}}\mbox{,} \quad \vartheta_{y} = \frac{V_{\mathcal{C}}}{Z_{0}}\mbox{,} \quad \vartheta_{z} = \frac{W_{\mathcal{C}}}{Z_{0}}
\end{aligned}
\end{equation}
Then, according to the derivations in \cite{Croon2013a}, substituting \cref{eqobs:4} and \cref{eqobs:5} into \cref{eqobs:1} leads to the following expressions for translational optical flow:
\begin{equation}\label{eqobs:6}
\begin{aligned}
u_{T} &= (-\vartheta_{x} + \vartheta_{z}x)(1-Z_{X}x-Z_{Y}y)\\
v_{T} &= (-\vartheta_{y} + \vartheta_{z}y)(1-Z_{X}x-Z_{Y}y)\\
\end{aligned}
\end{equation}
From \cref{eqobs:6}, and under the aforementioned assumptions, the scaled velocities, which provide non-metric information on camera ego-motion, can be derived from the translational optical flow of multiple image points. $\vartheta_{x}$ and $\vartheta_{y}$ are the opposites of the so-called \textit{ventral flows}, a quantification of the average flows in the $X$- and $Y$-axis of $\mathcal{C}$ respectively \cite{PijnackerHordijk2018}. Hence, $\omega_{x} = -\vartheta_{x}$ and $\omega_{y} = -\vartheta_{y}$. On the other hand, $\vartheta_{z}$ is proportional to the \textit{divergence} of the optical flow field, $D = 2\vartheta_{z}$ \cite{PijnackerHordijk2018}.

The flow divergence can alternatively be estimated simply by the relative change in the distance ($l$) between any two points at over time (t) \cite{Ho2018}. This method is referred to as \emph{size divergence} ($D_{s_{t}}$). A reliable estimate of the divergence ($\hat{D}$) can be generated by averaging the divergence estimate from a set of $N$ points in the image.
\begin{equation}
\begin{aligned}
    D_{s_{t}} &= \frac{1}{\Delta t}\frac{l_{t-\Delta t} -l_{t}}{l_{t-\Delta t}} \\
    \hat{D} &= \frac{1}{N}\sum_{i=1}^{N}D_{s_{t}}
\end{aligned}    
\end{equation}

In this work, we used a FAST corner detector and a Lukas--Kanade tracker implemented using OpenCV as in \cite{Ho2018}, we refer the reader there for more details. Throughout this paper, $N$ is limited to 100 if there were more than 100 points or simply all points if fewer have been tracked.


\section{Flight Platform and Simulation Environment} \label{sec:platform}
\begin{figure}[pos=!tb]
\centering
\includegraphics[width=0.6\linewidth]{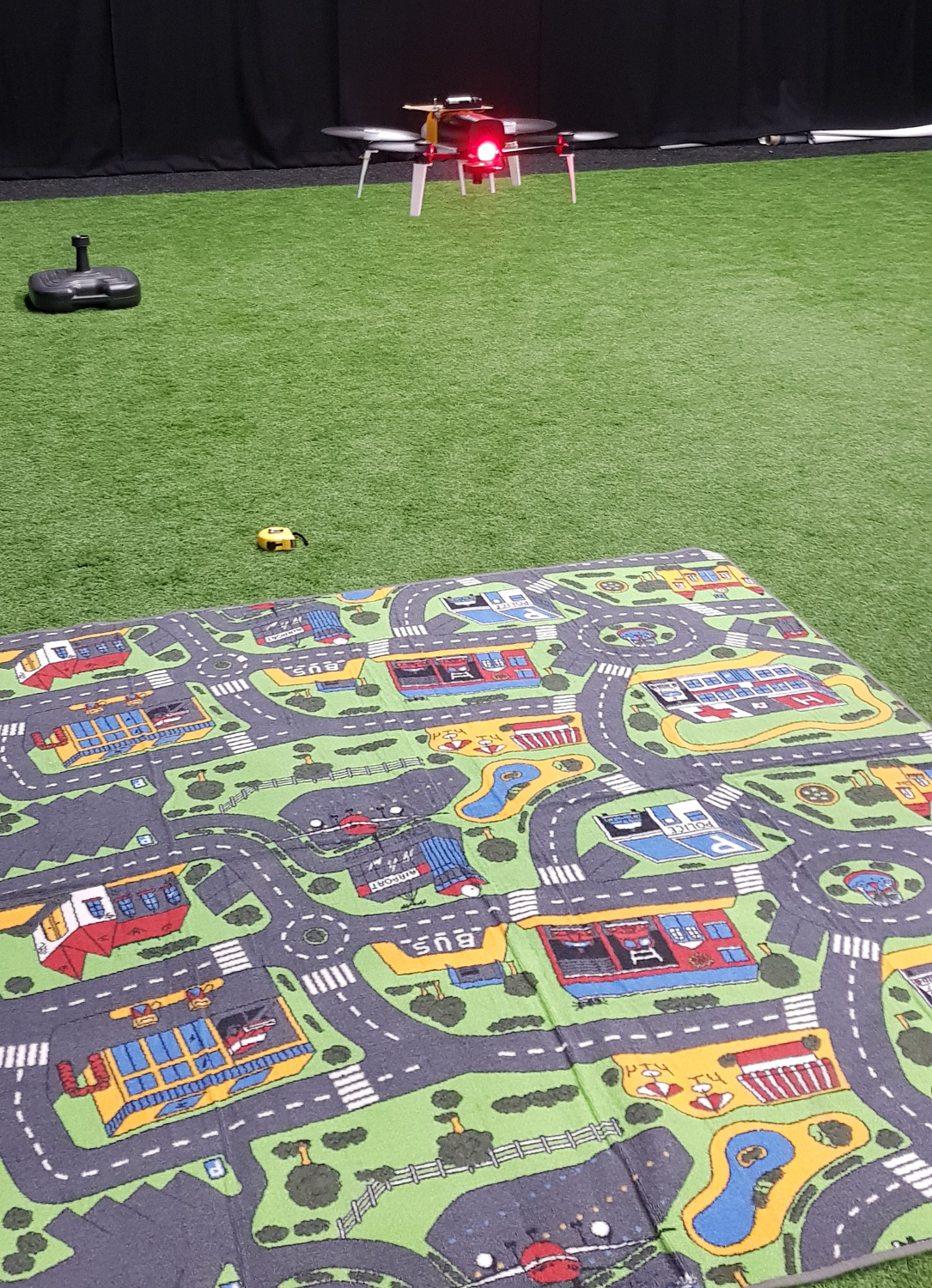}
\caption{Parrot Bebop 2 quadrotor MAV equipped with SEEM1 Dynamic Vision Sensor.}
\label{fig:bebop}
\end{figure}

The flight platform used in this paper is the Parrot Bebop 2 quadrotor MAV\footnote{\url{https://www.parrot.com/nl/en/drones/parrot-bebop-2}}, a picture of this vehicle flying in our indoor test environment can be found in \cref{fig:bebop}. This vehicle is equipped with a 780 MHz dual-core Arm Cortex A9 processor, forward and downward facing CMOS cameras, sonar, and barometer enabling autonomous flight capabilities for up to 25 minutes. Full 3D flight control is enabled with the onboard use of the open source PaparazziUAV autopilot software \cite{Hattenberger2014}. In this work, we extract global optical flow from the downward facing camera using the Lucas--Kanade optical flow method executed onboard the vehicle \cite{Lucas1981}\footnote{Code used in this paper can be found \url{https://github.com/kirkscheper/paparazzi/tree/updated_event_based_flow}}.

Throughout the paper, an Optitrack\footnote{https://optitrack.com/} motion capture system was used to measure a ground truth of the vehicle position and motion. As the control task in this paper is only in the vertical axis, this ground truth measure was communicated to the vehicle to facilitate the control of the lateral axis of the vehicle. This was however not used in the vertical loop where instead the optical flow estimated onboard the vehicle was used.

To optimize our behavior in simulation, we first need a model of the vehicle. To this end we use a simple dynamical model of the vehicle, which is restricted to vertical motion only. The thrust generated by the rotors is modeled as a first order response with the dynamics defined in \eqref{eq:thrust_dyn}.
\begin{equation} \label{eq:thrust_dyn}
(\Delta t + \tau_T)\dot{T}_i = T_{sp} - T_{i-1}
\end{equation}
where $T_{sp}$ is the thrust set-point and spin-up spin-down time constant $\tau_T$ has a nominal value of 0.02 s. The thrust output is limited in the range $[$-0.8$\cdot$g, 0.5$\cdot$g$]$ as a conservative model the maximum acceleration of the real vehicle.

The model used to describe the divergence estimation is based on the work presented in \cite{Ho2016}. The observed divergence is the result of adding latency to the true divergence along with two types of noise, simple white noise drawn from $N$(0,$\sigma_{w}^2$) and an additional noise proportional to the divergence magnitude drawn from $N$(0,$\sigma_{p}^2$). \cite{Ho2016} identified typical values for $\sigma_{w}$ and $\sigma_{p}$ as 0.1 s$^{-1}$ and the latency $L$ in the range of [50, 100] ms. We use similar nominal values for the standard deviations $\sigma_{w}$ and $\sigma_{p}$ are 0.1 s$^{-1}$ and an exaggerated range for the latency $L$ in the range of [1,4] samples or [20, 133] ms. Some additional computational jitter has been included, this simulates the situation of missed frames which happens when there are either insufficient or too many image features to be tracked. The chance of a missed frame is randomly determined with a given probability held constant for each simulation run randomly drawn from the range [0, 0.2].

%
%
\section{Baseline Performance}
\label{sec:baseline}

Before we start to optimize a divergence based landing controller, let us first investigate the naive approach of a constant gain, constant divergence landing as studied in \cite{Herisse2012,VanBreugel2014,Croon2016}.

\begin{figure}[pos=!tb]
\centering
\includegraphics[width=0.45\textwidth]{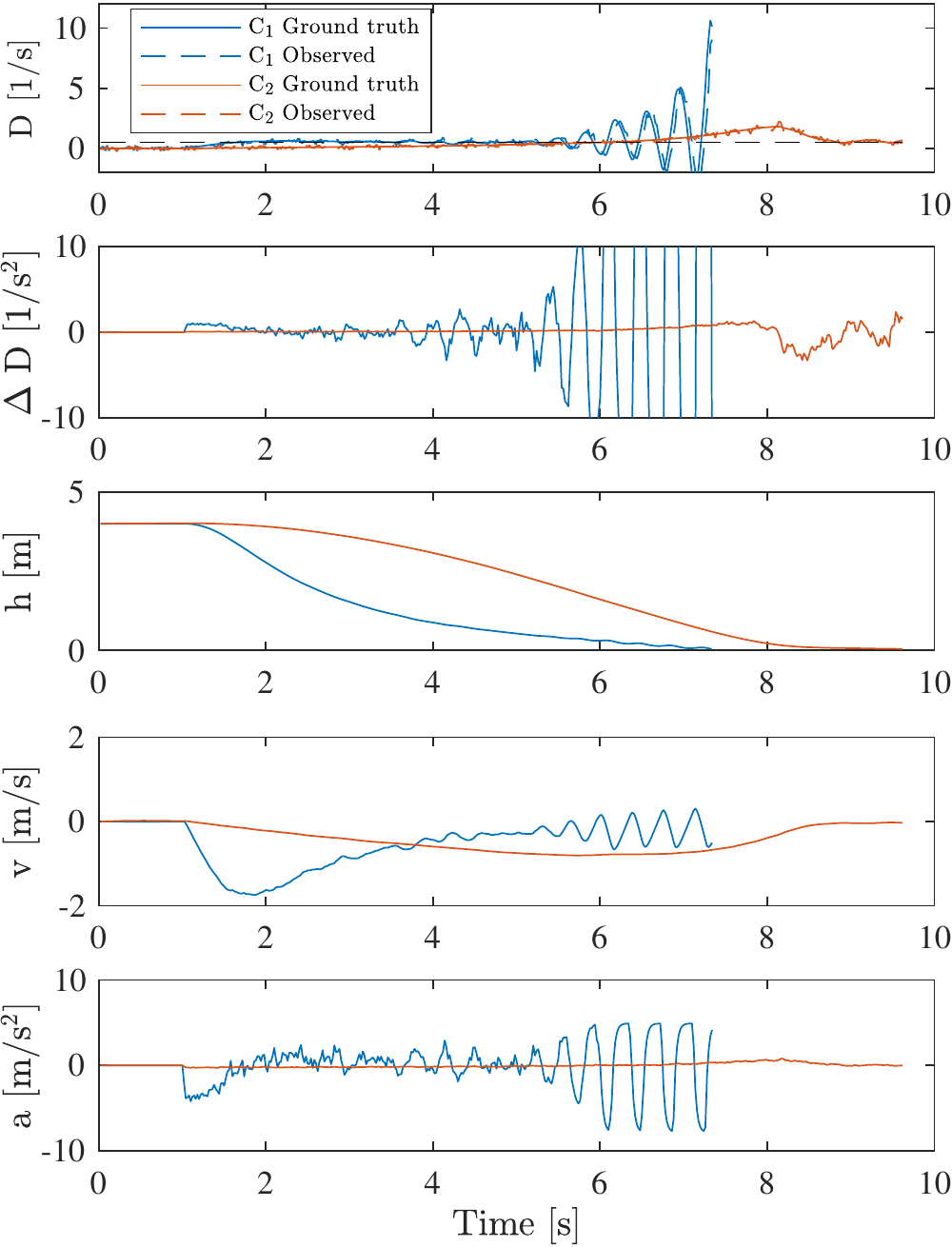}%
\caption{Time history of a constant divergence landing with two controllers of differing gain. $C_1$ is a controller with high gain and $C_2$ with low gain.}
\label{fig:hand_designed}
\end{figure}

\begin{figure}[pos=H]
\centering
\subfloat[Controller $C_1$]{\includegraphics[width=0.48\textwidth]{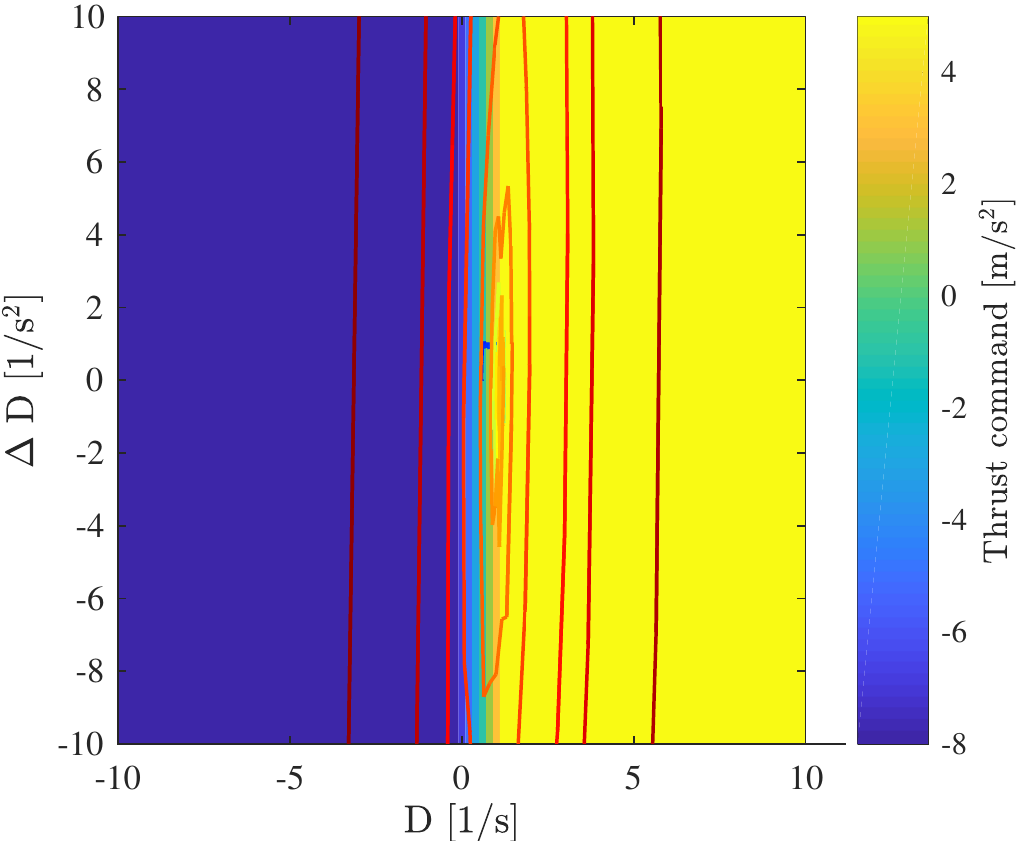}%
\label{fig:ss_mapping_c1}}
\\
\subfloat[Controller $C_2$]{\includegraphics[width=0.48\textwidth]{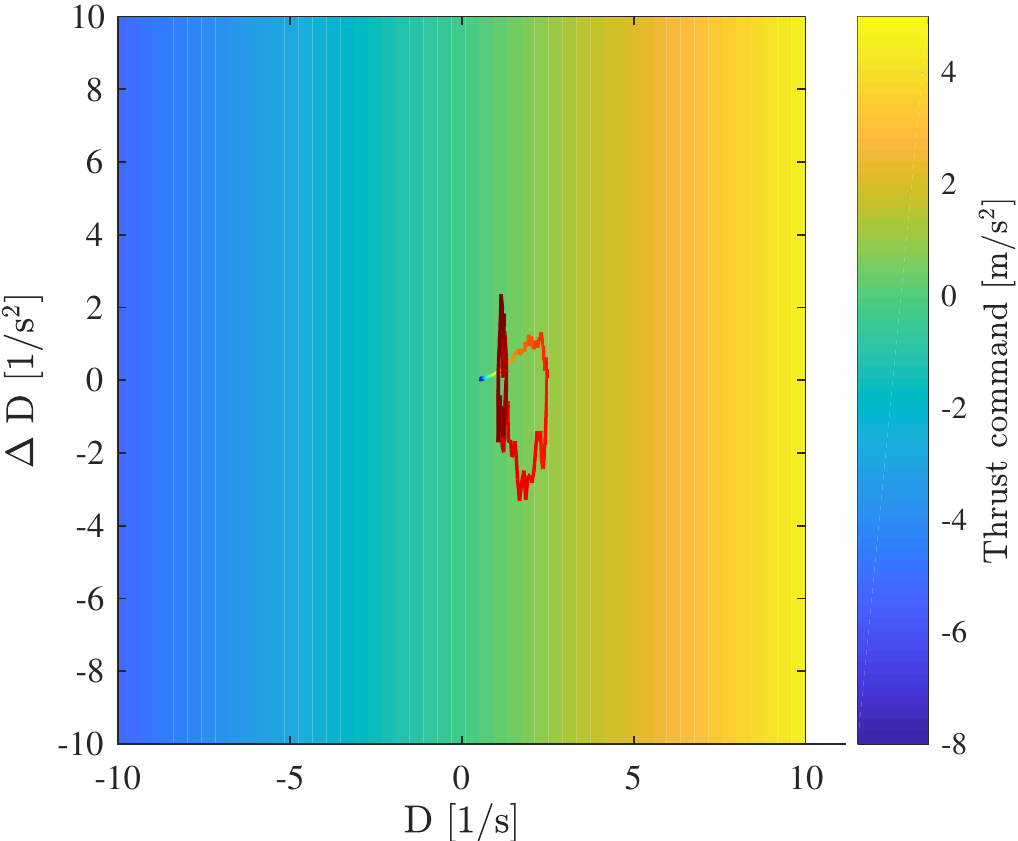}%
\label{fig:ss_mapping_c2}}
\caption{Steady state input-output mapping for hand designed controllers.}
\label{fig:ss_mapping_hand_designed}
\end{figure}

\cref{fig:hand_designed} shows the time history of a two constant divergence landing controllers performing a landing with a divergence set-point of 0.5. Controller $C_1$ has a relatively high gain and $C_2$ a low gain. The controllers are activated 1 s after the simulation starts. \cref{fig:ss_mapping_hand_designed} shows the steady-state response of the controller with the time history control signal super-imposed. 

These plots show that controller $C_1$ quickly reaches the desired divergence but as the vehicle descends the controller becomes increasingly unstable. This instability is due to delays in the sensing and control loop and is well described in \cite{Croon2016}. In contrast, $C_2$ does not quickly achieve the desired divergence due to the low gain in the controller. Additionally, due to the non-linear relationship between the divergence and acceleration, the vehicle overshoots the desired set-point before finally slowing. Despite this poor tracking performance, the low gain does delay the onset of instability observed with $C_1$, resulting in a rather smooth landing profile.


%
%
\section{Evolutionary Optimization}
\label{sec:evo}
Every evolutionary process is defined by: a population of candidate individuals, each with a given policy which must be evaluated; a way to evaluate these policies; a selection mechanism to filter out bad policies; and a method to alter the individuals to generate new policies. Here, we use a mutation-only evolutionary algorithm similar to ($\mu$ + $\lambda$) approach, where a population is maintained from which offspring are generated using a mutation operator. Offspring that are better than members of the population replace these members. The difference with the standard implementation is that we retest the current population on a random set of simulation parameters with every generation, rather than once as is commonly done. With the high level of non-determinism in our simulated environment, this ensures no individuals are preferred simply because they were tested on an \emph{easy} set of conditions.

The evaluation of the individuals is done by simulating the policy on four independent simulation runs, initializing the vehicle at a standstill from four different altitudes, namely 2, 4, 6 and 8 m. The simulation ends when the vehicle exceeds 15 m above the ground, gets within 5 cm of the ground or exceeds 30 s of simulation time. 

We use a multi-objective approach here, where the individuals must minimize three fitness functions measured at the end of a simulation: the total time to land ($f_1$); the final height ($f_2$); and the final velocity ($f_3$). NSGA-II is used to perform a non-dominated sorting of the population and determine which individuals are better than others in this multi-objective framework \cite{Deb2002}. 

All the simulation parameters mentioned in \cref{sec:platform} are randomly perturbed and set at the start of a generation. The ranges of the evolutionary parameters used in this work are summarized in \cref{tab:evo}. Altogether, this should encourage the optimization to develop a policy that can reliably land quickly and safely from different altitudes. This is all developed using the DEAP framework \cite{DEAP_JMLR2012} with multi-threaded implementation utilizing scoop in Python\footnote{The software is openly available at: \url{https://github.com/DEAP/deap}}\footnote{Software used for the evolutionary process in this paper is openly available at: \url{https://github.com/kirkscheper/divergence_landing}}.

\begin{table}
\caption{Evolutionary parameters}
\label{tab:evo}
\centering
\begin{tabular}{l|l}
Name & Value\\
\hline
Number of Generations & 250 \\
Number of Runs & 4 \\
\hline
Range of delays & [1, 4] samples \\
Range of computational jitter probability & [0, 0.2] \\
Range of divergence noise ($\sigma_{w}^{2}$) & [0.05, 0.15] 1/s\\
Range of divergence noise ($\sigma_{p}^{2}$) & [0, 0.25] 1/s\\
Range of thrust time constant ($\tau{T}$) & [0.005, 0.04] \\
Range of simulation frequency & [30, 50] Hz \\
\end{tabular}
\end{table}

The policy of each individual is encoded in a simple neural network. The neural potential ($\gamma$) of each neuron in the neurocontroller is updated with a simple discrete time Euler integration as described in \cref{eq:integration}.
\begin{equation} \label{eq:integration}
\gamma(t) = \gamma(t-1) + \dot\gamma(t-1) \Delta t
\end{equation}
\noindent{where $t$ represents the current time step and $\Delta t$ is the time step of the integration.}

The input to the neurocontroller is the simulated divergence and the derivative of the divergence $\Delta D = \frac{D_t - D_{t-\Delta T}}{\Delta t}$. The output was used to control the thrust of the vehicle leading to an acceleration. We utilized three types of neural networks to investigate the effect of recurrent connections on the evolved solution. We implemented a feed-forward neural network (hereafter referred simply as NN), a recurrent neural network (RNN) and a continuous time recurrent neural network (CTRNN). All networks had three layers, the first with 2 neurons, the hidden layer with 8 neurons and the output with 1 neuron.

\subsection{NN}
The neural potential of an NN is defined by the instantaneous inputs to the network such that the potential of a neuron $i$ in a given layer $l$ connected to $N^{l-1}$ neurons in the previous layer is determined simply by:
\begin{equation}
\gamma^l_i = \sigma^l(\sum_{j=1}^{N^{l-1}}w^l_{ij}\gamma^{l-1}_j) + \theta^l_i + I^l_i
\end{equation}

\noindent{where $w_{ij}$ is the weight of the neural connection between the neuron $i$ in layer $l=(2,3,\cdots)$ and a given neuron $j$ is the previous layer, $\theta$ is the neural bias, $I$ is the external input to the neuron and $\sigma$ is the activation function of the neuron. Here, the activation function is linear for the output layer and the Rectified Linear Unit (ReLU) function for the hidden layer. We can rewrite this equation using \cref{eq:integration} to generate the neural dynamics.}
\begin{equation}
\Delta t \dot\gamma_i = -\gamma_i + \sigma(\sum_{j=1}^{N}w_{ij}\gamma_j) + \theta_i + I_i
\end{equation}

\subsection{RNN}
The NN has no effective way of explicitly considering previous states in determining it's current action. The behavior is simply a result of the emergent interaction of the vehicle actions and the environment. Adding explicit memory may enable the vehicle to exhibit more complex behaviors. RNNs are not only affected by an external input and the weighted sum of connected neurons but also by a weighted internal connection to the previous potential as shown in \cref{eq:rnn_update}.
\begin{equation}\label{eq:rnn_update}
\dot\gamma_i = \gamma_i r_{i} + \sigma(\sum_{j=1}^{N}w_{ij}\gamma_j) + \theta_i + I_i
\end{equation}
\noindent{where $r$ is the weight applied to the recurrent connection. Like the NN, the activation function is linear for the output layer and the ReLU function for the hidden layer.}

Again, using \cref{eq:integration}, we can derive the neural dynamics:
\begin{equation}
\Delta t \dot\gamma_i = \gamma_i(r_{i} - 1) + \sigma(\sum_{j=1}^{N}w_{ij}\gamma_j) + \theta_i + I_i
\end{equation}

\subsection{CTRNN}
One pitfall of the RNN is that the recurrent connection does not consider the time between updates. This can be an important consideration for systems with variable time steps between updates. As such, we have also implemented the classic CTRNN as shown in \cref{eq:CTRNN} \cite{Beer1995}.
\begin{equation}\label{eq:CTRNN}
(\Delta t + \tau_i)\dot\gamma_i = -\gamma_i + \sum_{j=1}^{N}w_{ij}\sigma(\gamma_j + \theta_j) + I_i
\end{equation}
\noindent{where $\tau$ is its time constant ($\tau > 0$). The hyperbolic tangent activation function is used here ($\sigma(x)$ = tanh = $(e^{x} - e^{-x})/(e^{x} + e^{-x})$).}

\section{Evolution Results} \label{sec:evo_results}
\begin{figure}[pos=!tb]
\centering
\includegraphics[width=0.39\textwidth]{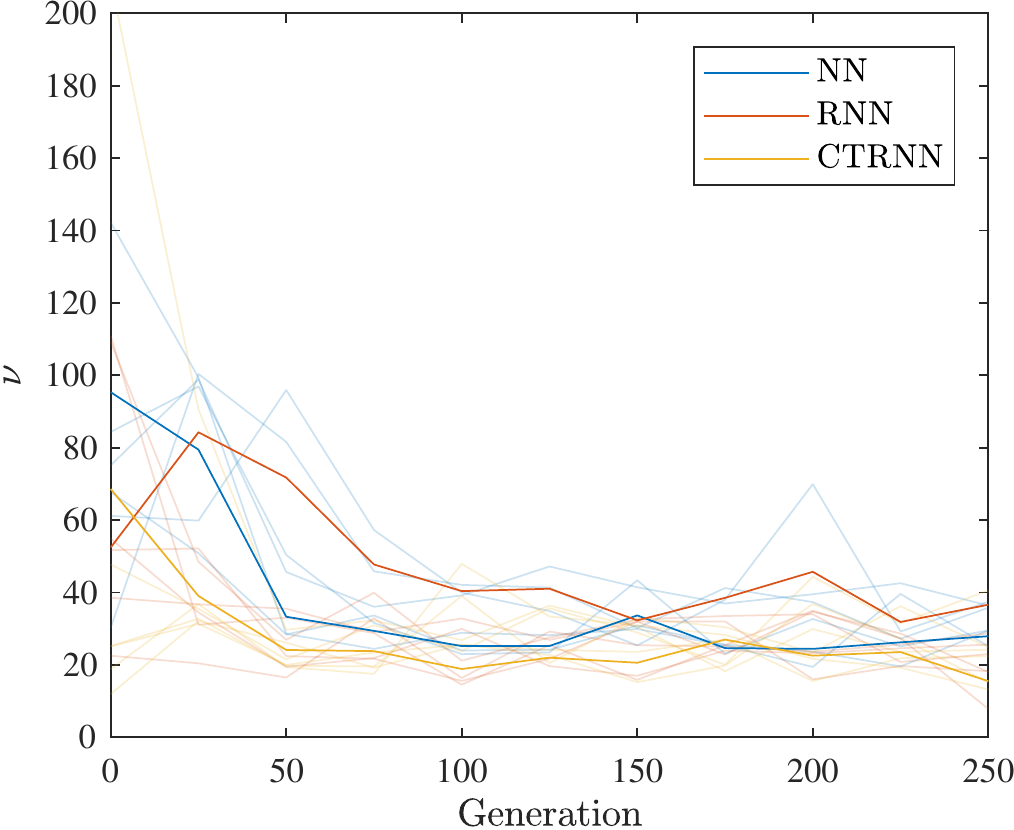}
\caption{Performance of genetic optimization measured with $\nu$ which is the ratio of the encapsulated volume and area of the pareto front. A smaller $\nu$ would suggest a general improvement in the minimization of all individuals while they spread out over the available optimum policies. The pareto front was generated by evaluating the entire population at each generation to the same set of simulated environmental conditions.}
\label{fig:evo_performance}
\end{figure}

The use of the non-dominated sorting and selection, results in the genetic optimization spreading the population of policies over the pareto front of the fitness landscape. As such, to evaluate the performance of the optimization, it is sensible to look at the ratio of the encapsulated volume and the area of the pareto front ($\nu = volume/area$) as shown in \cref{fig:evo_performance}. This figure shows that the performance generally improves before leveling off after about 150 generations. This performance is also mostly stable as the performance remains flat after 150 generations. This also shows that this trend is consistent over the multiple initializations of the optimization and over the different types of neural architecture.

\begin{figure}[pos=!tb]
\centering
\includegraphics[width=.39\textwidth]{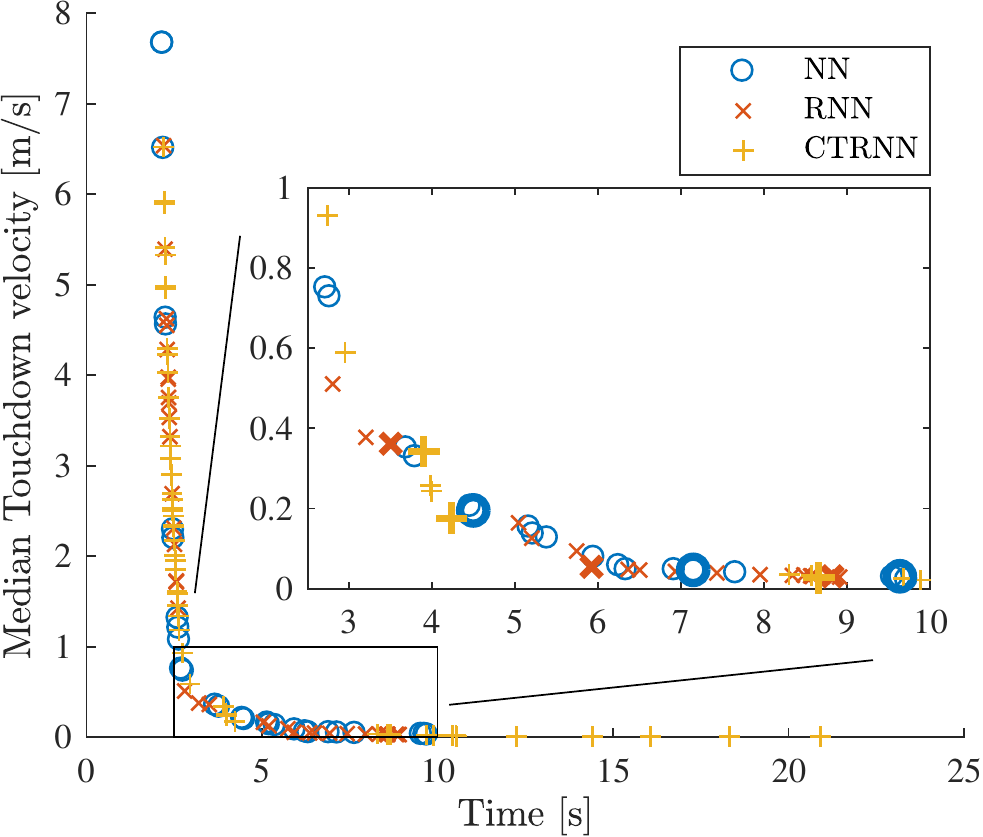}
\caption{Performance of pareto front tested on 250 evaluations. Individuals selected for further analysis are bold.}
\label{fig:pareto_front}
\end{figure}

\cref{fig:pareto_front} shows the accumulated Pareto front of all individuals from the different runs of the optimization. This figure only shows the performance on the touchdown velocity and the time fitness as the fitness based on the height was consistently minimized for all individuals. As such, all optimized policies converged to perform the desired landing task. This figure shows how the policies are spread over the inherent trade-off of reliable quick vs soft landing. The performance seems not to be strongly correlated with the neural architecture as all three types are represented in the pareto front. Additionally, the three architectures seem well distributed.

\begin{figure}[pos=!tb]
\centering
\includegraphics[width=.39\textwidth]{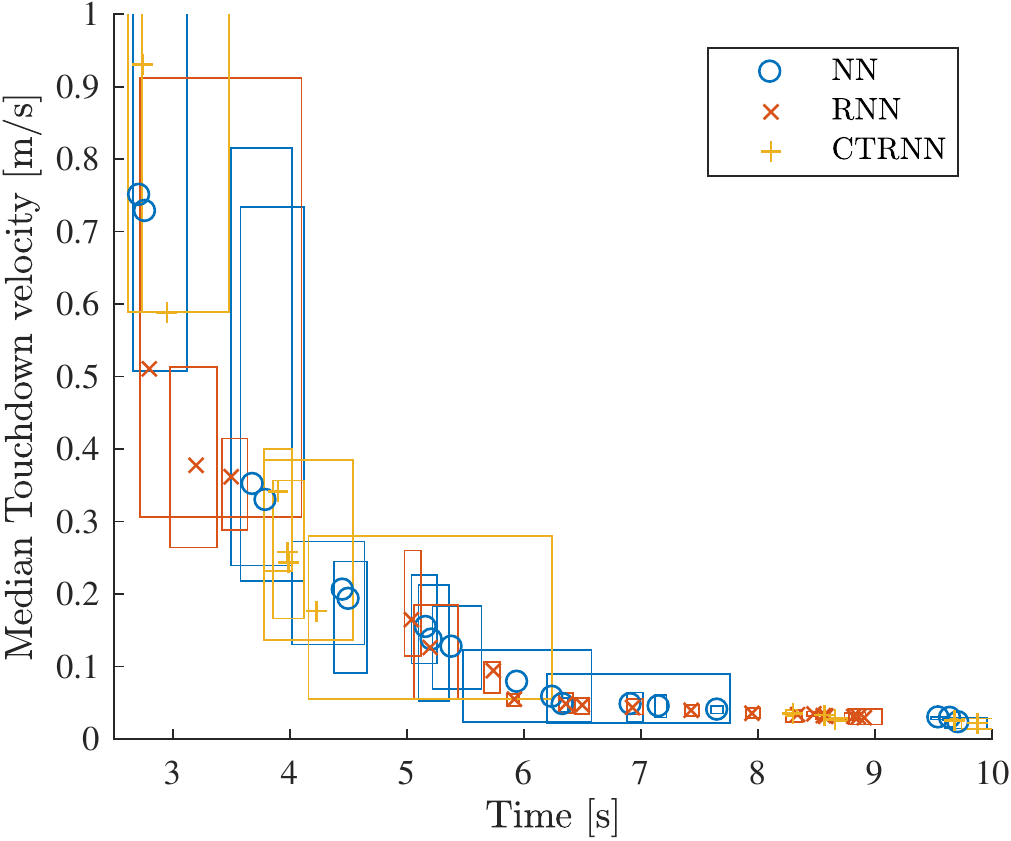}
\caption{Sensitivity analysis of a portion of pareto front showing 25$^{th}$, 50$^{th}$ and 75$^{th}$ percentile of performance over 250 evaluations.}
\label{fig:pareto_front_stats}
\end{figure}

To investigate the sensitivity of the performance to different environmental conditions, we subjected the individuals of the pareto front to a validation test with 250 simulations while varying the environmental settings of the run. All individuals where subjected to the same set of conditions to make for a fair comparison. \cref{fig:pareto_front_stats} shows that the individuals that optimized to have slow and soft landings have a small spread in the fitness performance whilst individuals that were optimized to faster landings have a larger variation in landing speed. This suggests that the policies that perform faster landings may show oscillatory behavior causing their touchdown speed to vary depending on the sensor noise or environmental perturbations. We will investigate this more below.

As it is not feasible to analyse them all, three individuals from the each architecture are be selected for further analysis of the landing behavior. The remainder of this section will dive deeper into the types of behaviors optimized by the different neurocontrollers.

\subsection{NN}
\begin{figure}[pos=!tb]
\centering
\includegraphics[width=0.45\textwidth]{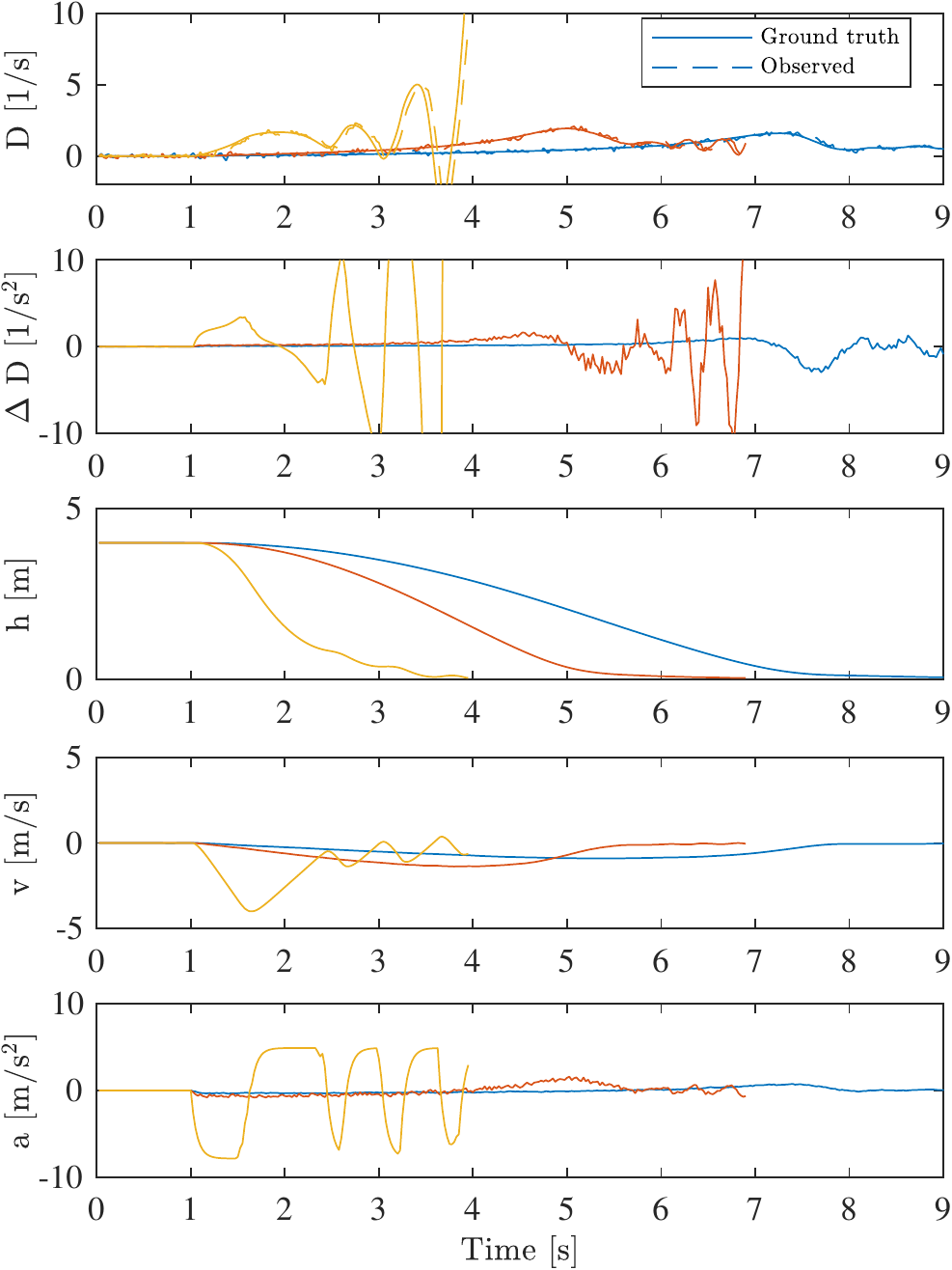}%
\caption{Vehicle states and observations from simulation with NN$_1$ (blue), NN$_2$ (red) and NN$_3$ (yellow).}
\label{fig:nn_sim}
\end{figure}

\begin{figure}[pos=H]
\centering
\subfloat[NN$_1$]{\includegraphics[width=0.48\textwidth]{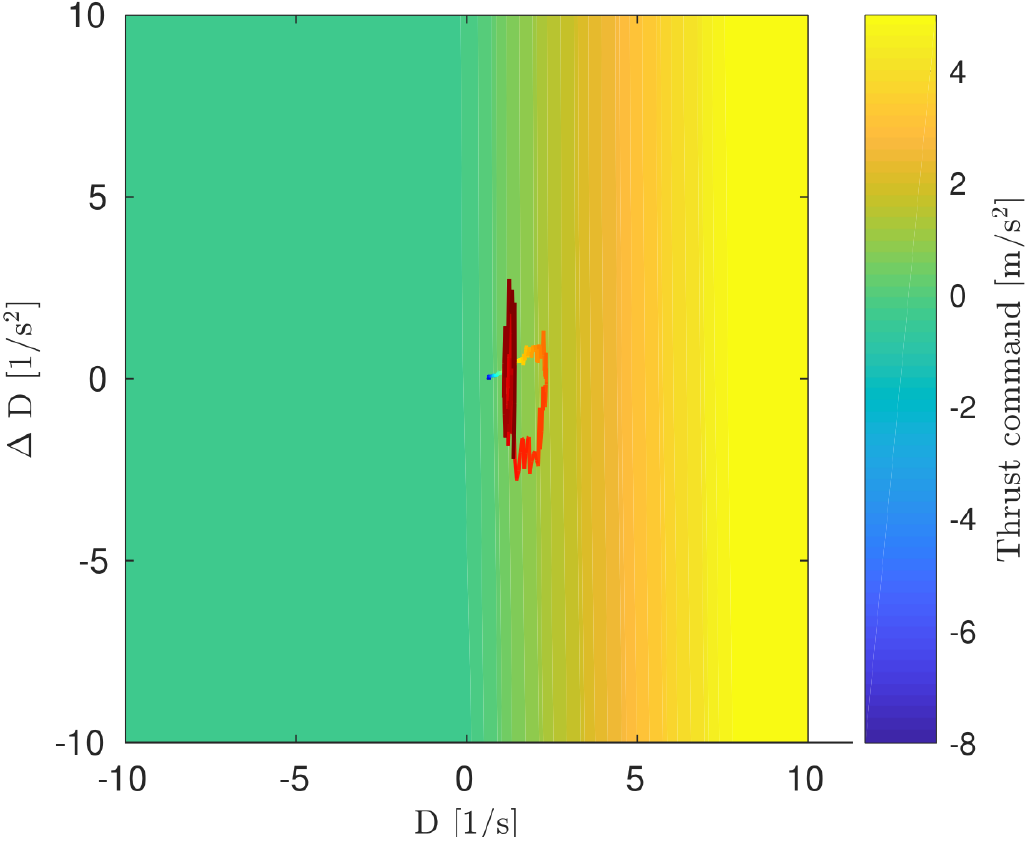}%
\label{fig:ss_mapping_nn_slow}}
\\
\subfloat[NN$_2$]{\includegraphics[width=0.48\textwidth]{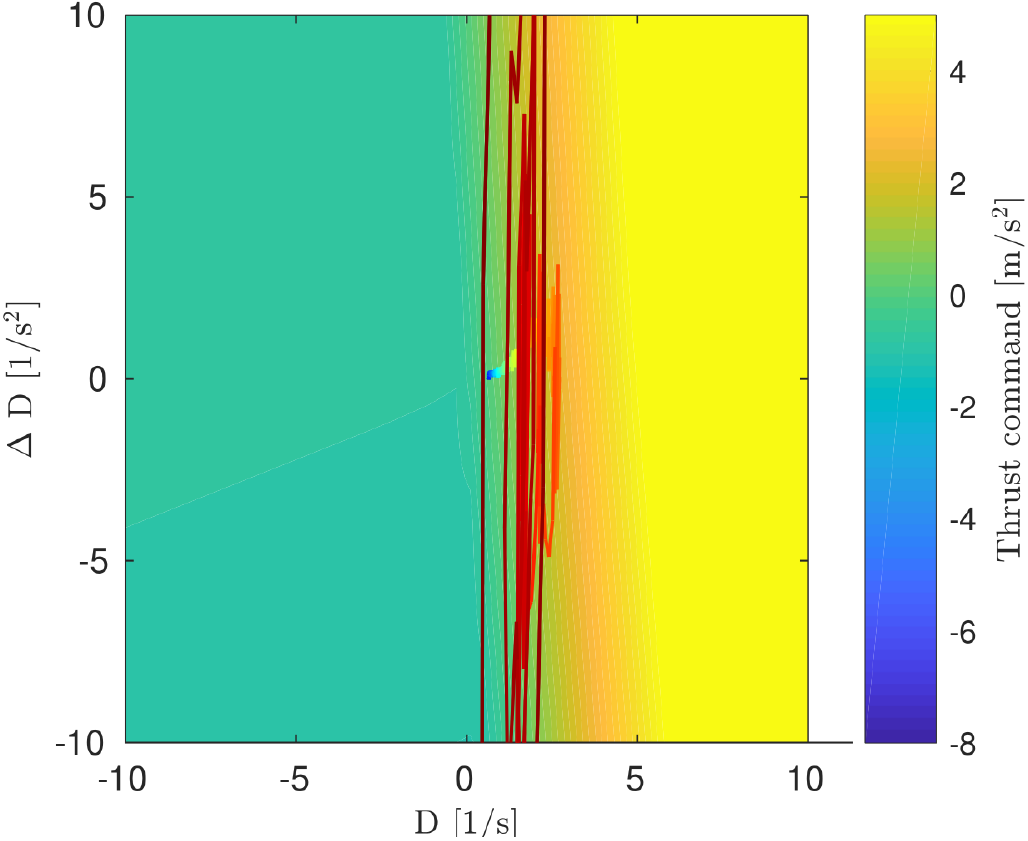}%
\label{fig:ss_mapping_nn_med}}
\\
\subfloat[NN$_3$]{\includegraphics[width=0.48\textwidth]{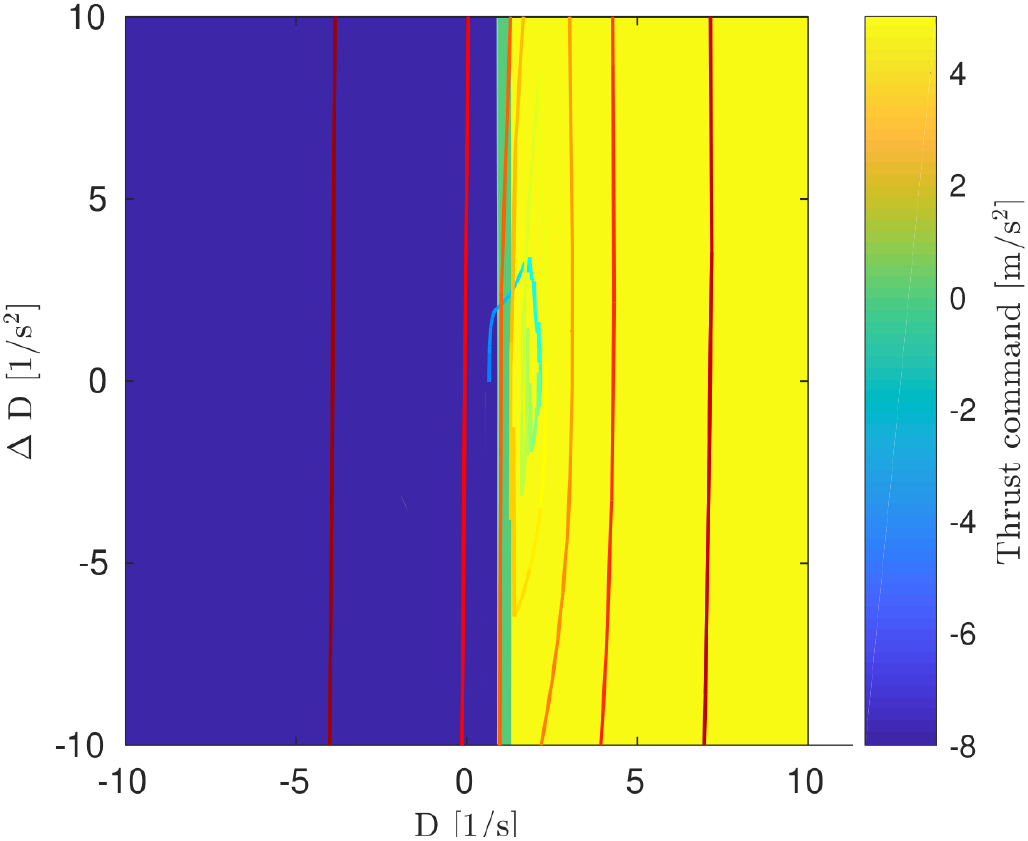}%
\label{fig:ss_mapping_nn_fast}}
\caption{Steady state input-output mapping for NN controllers.}
\label{fig:ss_mapping_nn}
\end{figure}

Three neurocontrollers were selected from the Pareto front for some further analysis, their performance is shown in \cref{fig:nn_sim} and their steady-state response in \cref{fig:ss_mapping_nn}. Controller NN$_1$ is a slow lander, NN$_3$ is a fast lander and NN$_2$ is intermediate. 

Looking first at the world states in \cref{fig:nn_sim}, we can see that NN$_1$ and NN$_2$ perform smooth landings with little oscillation and touchdown at very low velocities. Examining the steady-state response, NN$_1$ is similar to the low gain baseline C$_2$ except that instead of being a constant gain for all inputs, NN$_1$ is piece-wise linear with a high gain for positive values D and a low gain for negative values. This asymmetric control scheme is a significant result as this will delay the onset of the oscillation seen in higher gain controllers.

NN$_2$ is similar to NN$_1$ but has a higher gain and a noticeable gradient in the $\Delta$D with reduced thrust at negative $\Delta$D. This is also an interesting result as a negative $\Delta$D occurs when the vehicle is slowing while descending or accelerating while ascending. In both cases, it would indeed be desirable to reduce the control input.

NN$_3$ has clear oscillations and is similar to the high gain controller C$_1$ but is even higher gain, this causes the controller to act as a \emph{bang-bang} controller. This type of control will cause the rotors of the vehicle to try to spin up and down very often, however, due to their inertia, they do not achieve the desired values. As such this control scheme relies on the simulated spin-up and spin-down reaction time of the rotors ($\tau_{T}$) for the desired behavior to work well and will likely not transfer well to the real world. This may also explain why the controllers on the left of the pareto front in \cref{fig:pareto_front_stats} have vertically elongated bounding boxes.

\subsection{RNN}
\begin{figure}[pos=!tb]
\centering
\includegraphics[width=0.45\textwidth]{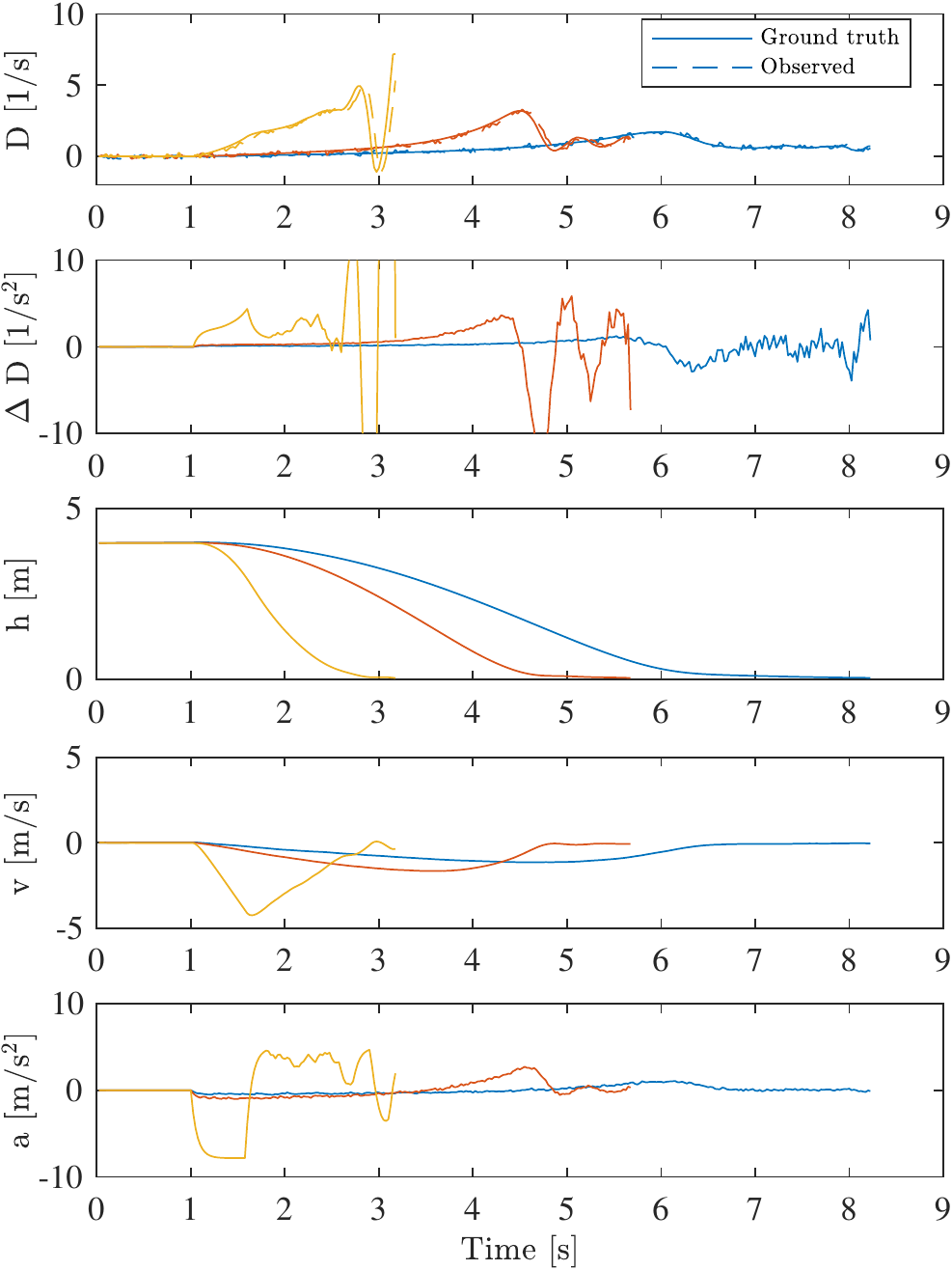}%
\caption{Vehicle states and observations from simulation with RNN$_1$ (blue), RNN$_2$ (red) and RNN$_3$ (yellow).}
\label{fig:rnn_sim}
\end{figure}

\begin{figure}[pos=H]
\centering
\subfloat[RNN$_1$]{\includegraphics[width=0.48\textwidth]{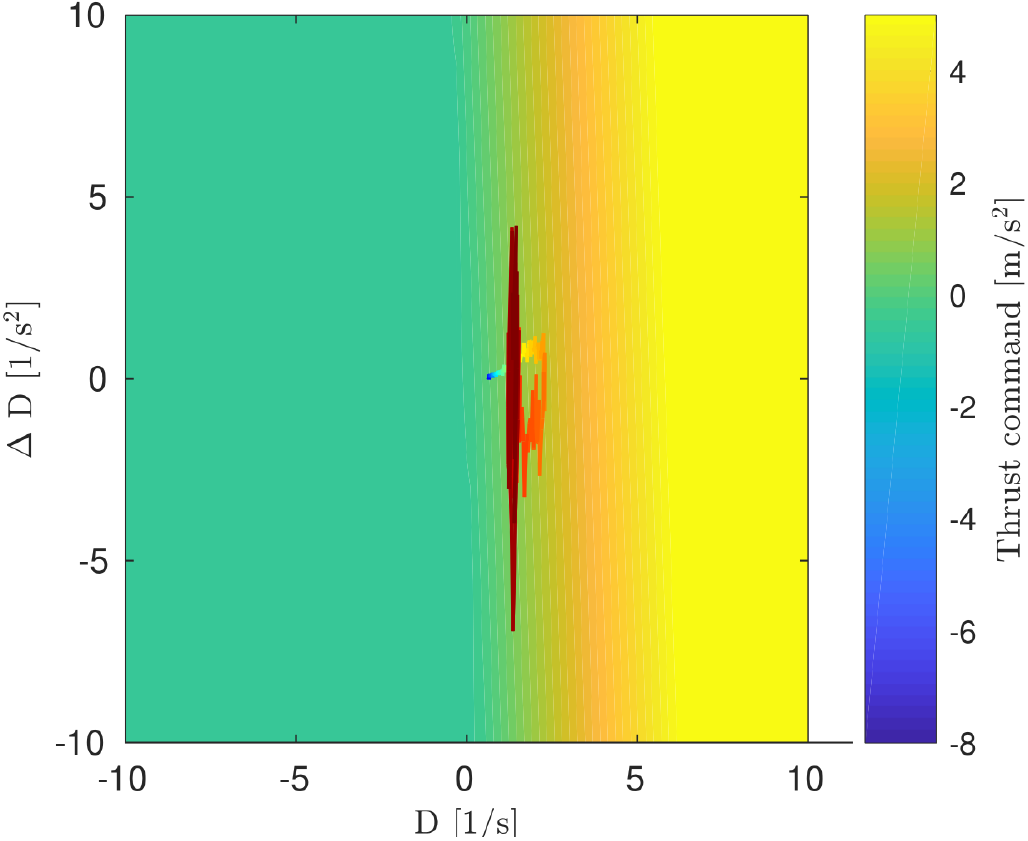}%
\label{fig:ss_mapping_rnn_slow}}
\\
\subfloat[RNN$_2$]{\includegraphics[width=0.48\textwidth]{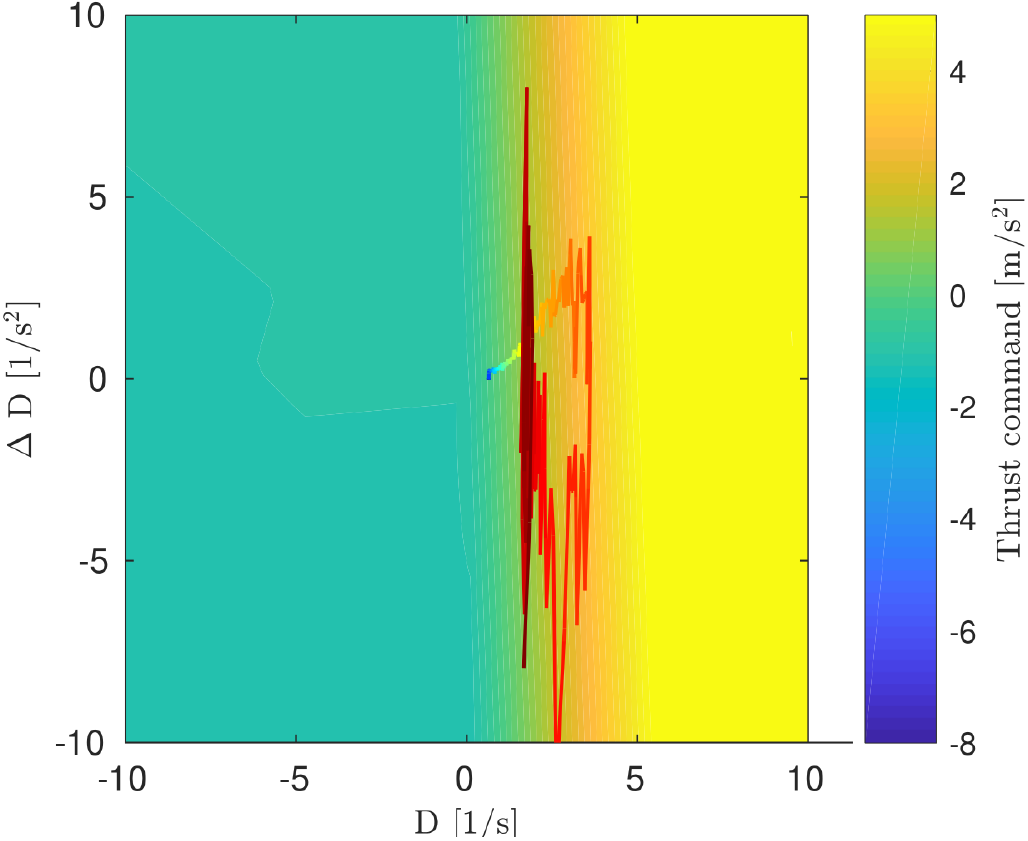}%
\label{fig:ss_mapping_rnn_med}}
\\
\subfloat[RNN$_3$]{\includegraphics[width=0.48\textwidth]{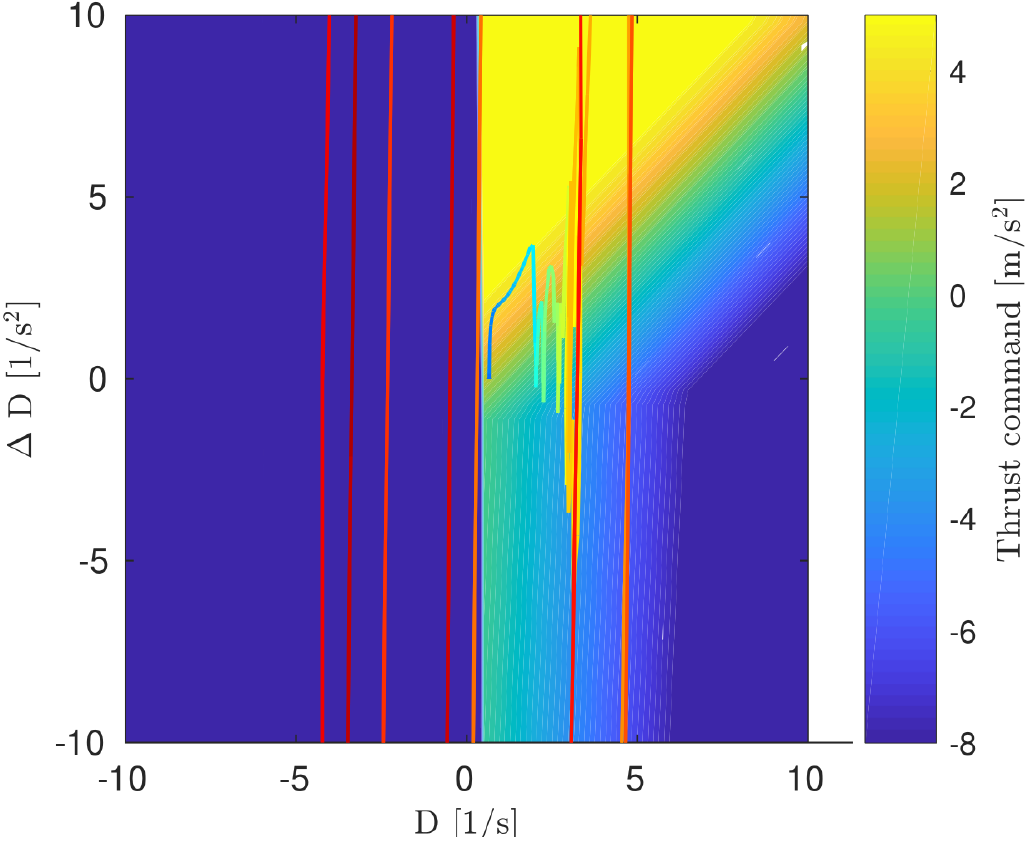}%
\label{fig:ss_mapping_rnn_fast}}
\caption{Steady state input-output mapping for NN controllers.}
\label{fig:ss_mapping_rnn}
\end{figure}

The selected RNN neurocontrollers are shown in \cref{fig:rnn_sim} and \cref{fig:ss_mapping_rnn}. RNN$_1$ and RNN$_2$ show similar behavior to the policies with NN$_1$ and NN$_2$. These relatively high gain landings with a gradient on the $\Delta$D seems to be a reliable way to perform this type of high speed yet smooth landing. RNN$_3$ is a little different than NN$_3$ in that high speed landings with a negative divergence rate have a lower throttle response. This would occur when the vehicle is descending quickly but slowing down. This control scheme would ensure that the vehicle doesn't over react when the vehicle is descending quickly but not accelerating towards the ground. This results in a reduced oscillatory behavior with the RNN$_3$ controller.

\subsection{CTRNN}
\begin{figure}[pos=!tb]
\centering
\includegraphics[width=0.45\textwidth]{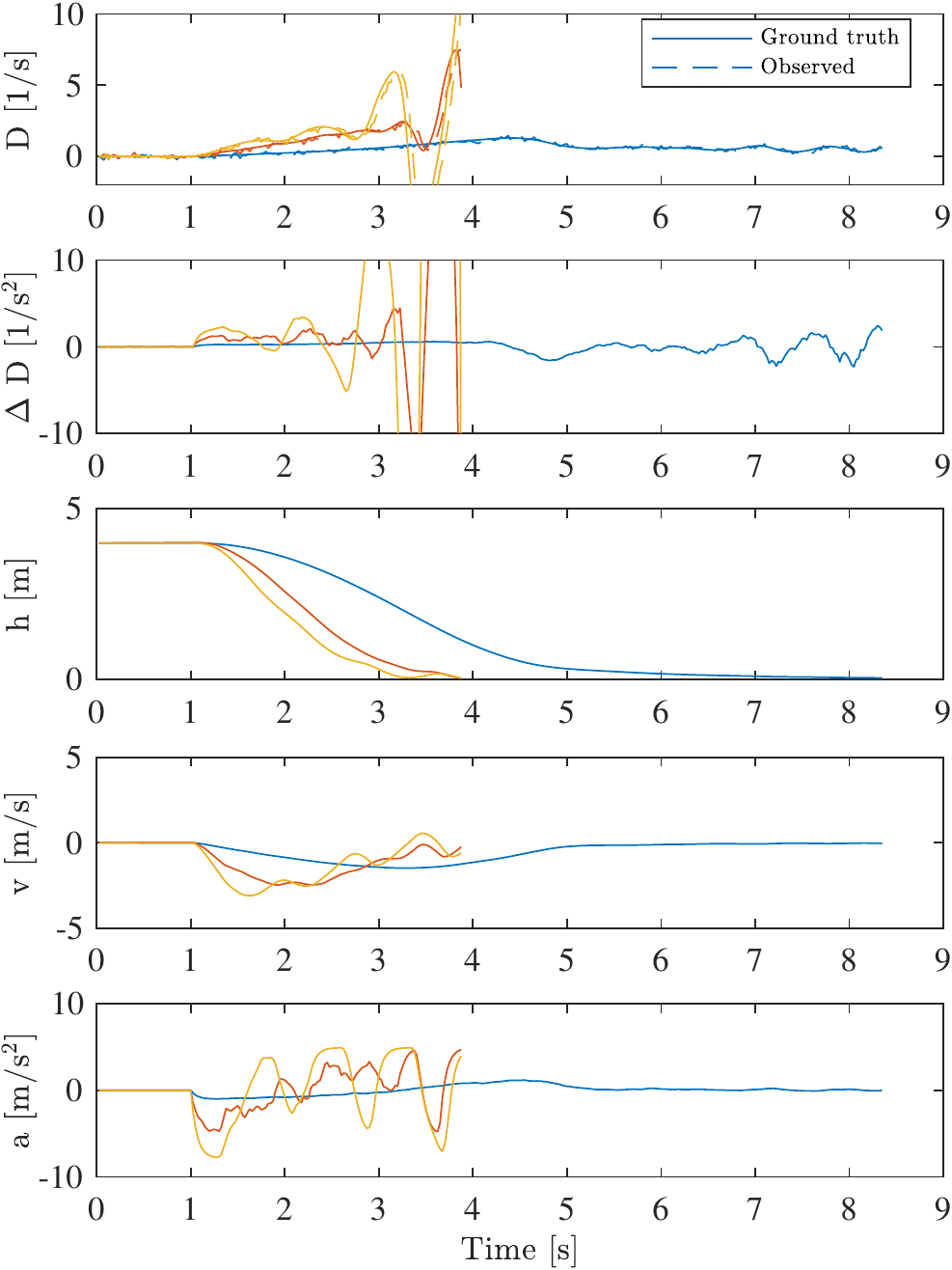}%
\caption{Vehicle states and observations from simulation with CTRNN$_1$ (blue), CTRNN$_2$ (red) and CTRNN$_3$ (yellow).}
\label{fig:ctrnn_sim}
\end{figure}

\begin{figure}[pos=H]
\centering
\subfloat[CTRNN$_1$]{\includegraphics[width=0.48\textwidth]{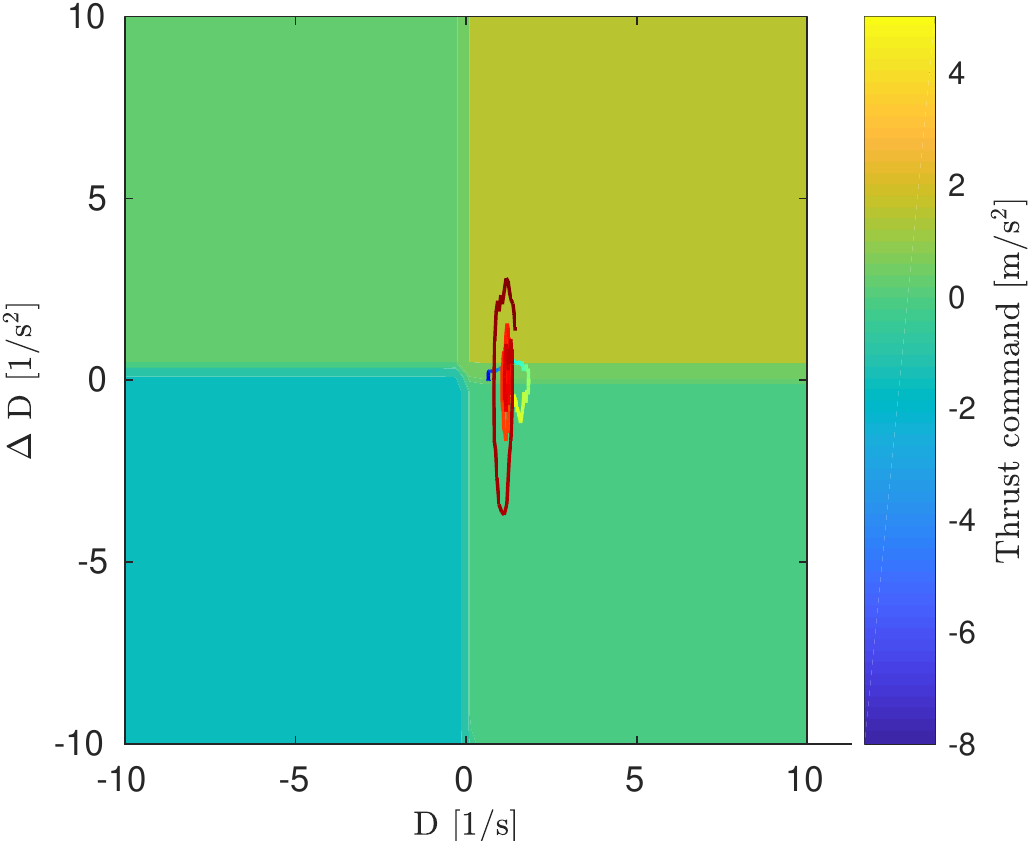}%
\label{fig:ss_mapping_ctrnn_slow}}
\\
\subfloat[CTRNN$_2$]{\includegraphics[width=0.48\textwidth]{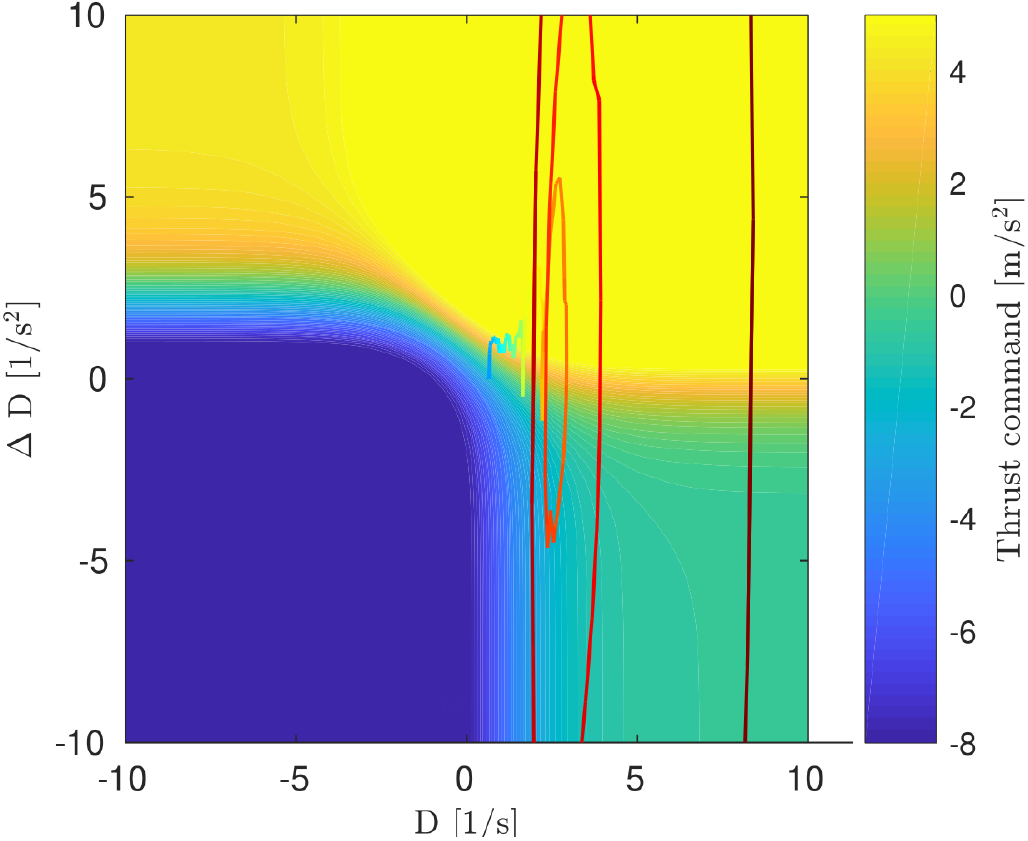}%
\label{fig:ss_mapping_ctrnn_med}}
\\
\subfloat[CTRNN$_3$]{\includegraphics[width=0.48\textwidth]{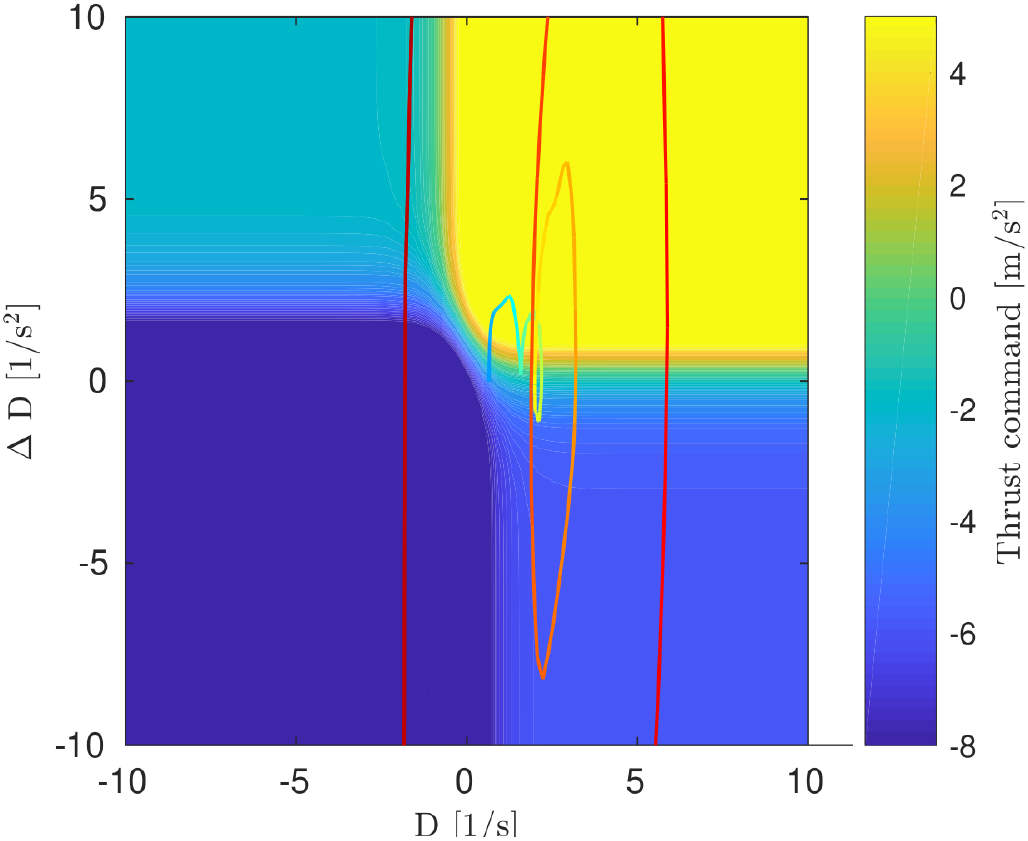}%
\label{fig:ss_mapping_ctrnn_fast}}
\caption{Steady state input-output mapping for NN controllers.}
\label{fig:ss_mapping_ctrnn}
\end{figure}

The three CTRNN neurocontrollers selected from the Pareto front in \cref{fig:ctrnn_sim} and \cref{fig:ss_mapping_ctrnn} all show variations on a similar control scheme. Effectively, these controllers split the control scheme into 4 segments, depending on the sign combination of the inputs D and $\Delta$D. When descending with an increasing divergence, the vehicle will try to decelerate. When descending with an decreasing divergence, the vehicle will decelerate less aggressively. When ascending with increasing rate, the vehicle will reduce thrust. Finally, when the vehicle is descending with a negative divergence rate, the vehicle will reduce thrust less aggressively than the increasing rate case. This results in high speed yet smooth landings with little oscillation. This approach is similar to that shown by RNN$_3$. Portions of these controllers seem discretized, but as they only show the steady state throttle response, the temporal response may be more smooth.

%
%
\section{Reality Gap}
\label{sec:realityGap}
Everything discussed so far has been in a simulated environment, one that is a significant simplification of reality to facilitate high speed evaluation of the neurocontrollers. As we move to the real world, we can therefore expect various differences leading to a reality gap. This section aims to identify some of these differences.

\begin{figure}[pos=!tb]
\centering
\includegraphics[width=0.45\textwidth]{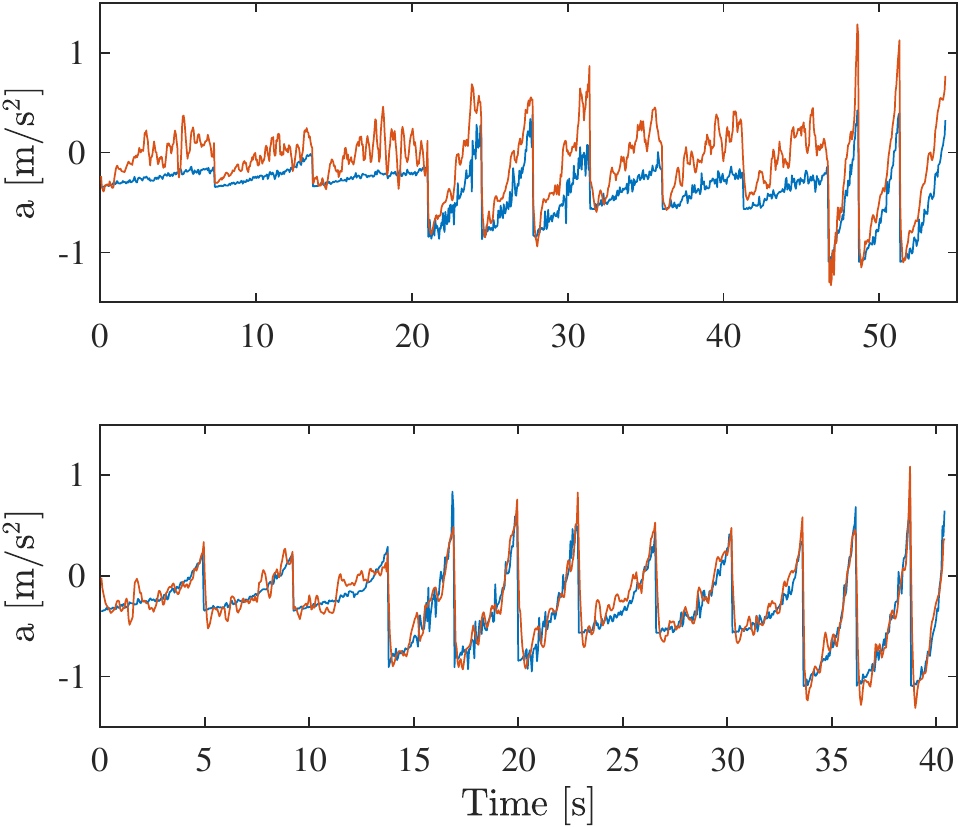}%
\caption{Thrust command (blue) and subsequent vehicle acceleration (red) for some real world landings. (Top)  Unmodeled dynamics such as drag and other non-linear aerodynamic effects lead to a poor thrust command tracking using a naive approach. (Bottom) The tracking error can be substantially reduced with the use of a simple closed loop PI controller.}
\label{fig:reality_gap_thrust}
\end{figure}

Let us first look at the control of the vehicle, in simulation, the output of the neurocontroller was acceleration, which after being fed through a low pass filter to simulate the spin-up of the rotors was implemented by the simulated rotors. In reality, this desired acceleration must first be converted to a thrust command for the rotors. If we use a naive approach here, we can simply determine a linear transform from desired acceleration to thrust, the results of which are shown in the top plot of \cref{fig:reality_gap_thrust}. This figure shows a set of real world neurocontroller landings using a constant scaling factor for desired acceleration to thrust. As the vehicle starts to move the command tracking is good but as it starts to descend the tracking degrades. This is almost certainly due, in part, to the unmodeled drag and non-linear aerodynamic effects of descending through the downwash of the propellers.

This poor tracking leads to a noticeable reality gap in the landing performance. This can be reduced with the use of a closed loop controller as proposed in \cite{Scheper2016a}, instead of a linear transform. The results using a Proportional-Integral (PI) controller, minimizing the error between the commanded and measured acceleration on the vehicle, are shown in the bottom plot of \cref{fig:reality_gap_thrust}. The controller effectively abstracts away from the raw motor commands to a desired acceleration, which significantly improves the tracking performance allowing us to cross this reality gap.


The goal of this paper is to highlight the use of abstracted inputs to improve the robustness of optimized policies to differences in the input. To do this we will investigate the performance of the vehicle with the use of two different types of cameras with significantly different divergence signal output performance characteristics.

\begin{figure}[pos=!tb]
\centering
\subfloat{\includegraphics[width=0.42\textwidth]{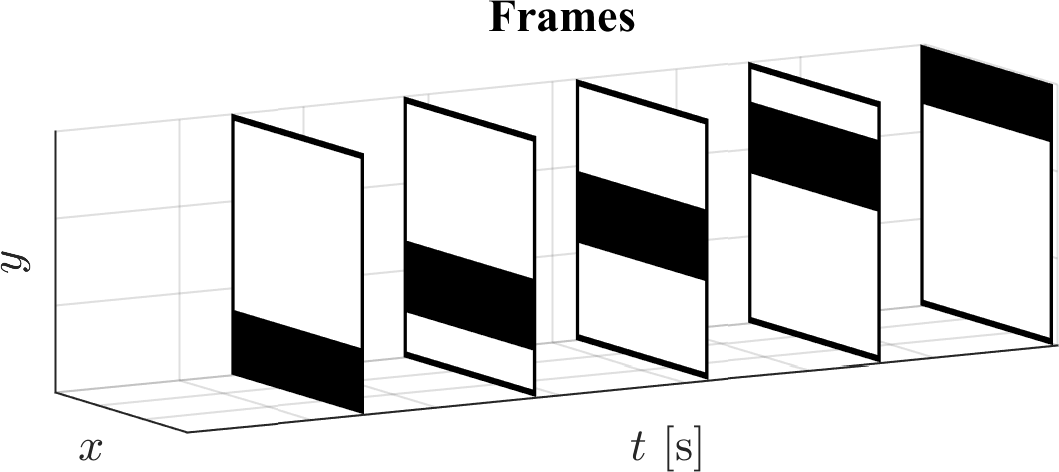}}\\
\subfloat{\includegraphics[width=0.42\textwidth]{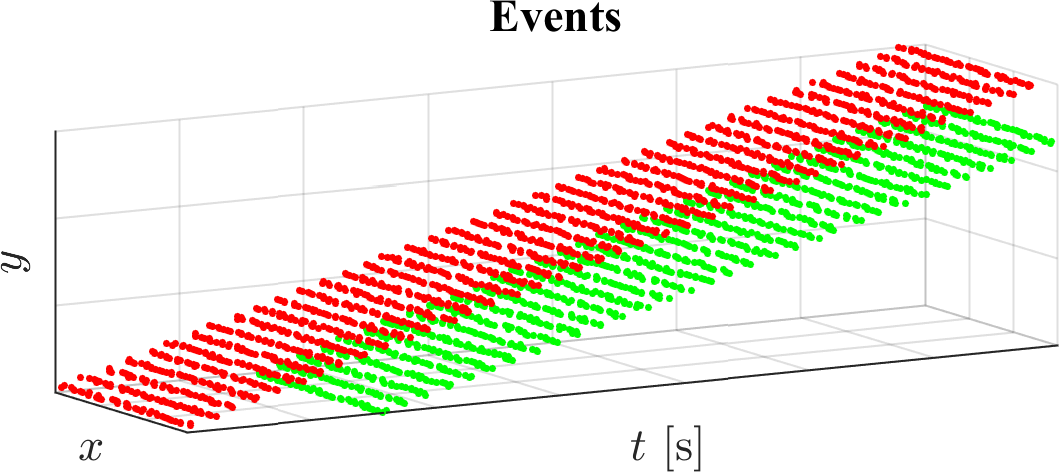}}
\caption{Depiction of the difference between frame-based and event-based vision data. Frame-based cameras measure light at all the pixels synchronously resulting in an image frame, optic flow is generated by identifying a feature in the current image in a previous image, the translation is the flow. Event-based cameras asynchronously generate an event for each pixel with a microsecond timestamp, optic flow is generated by fitting a plane through the 3D surface (x,y,t) of the most recent events.}
\label{fig:frameVSevent}
\end{figure}

\begin{figure*}\centering
\includegraphics[width=0.9\textwidth]{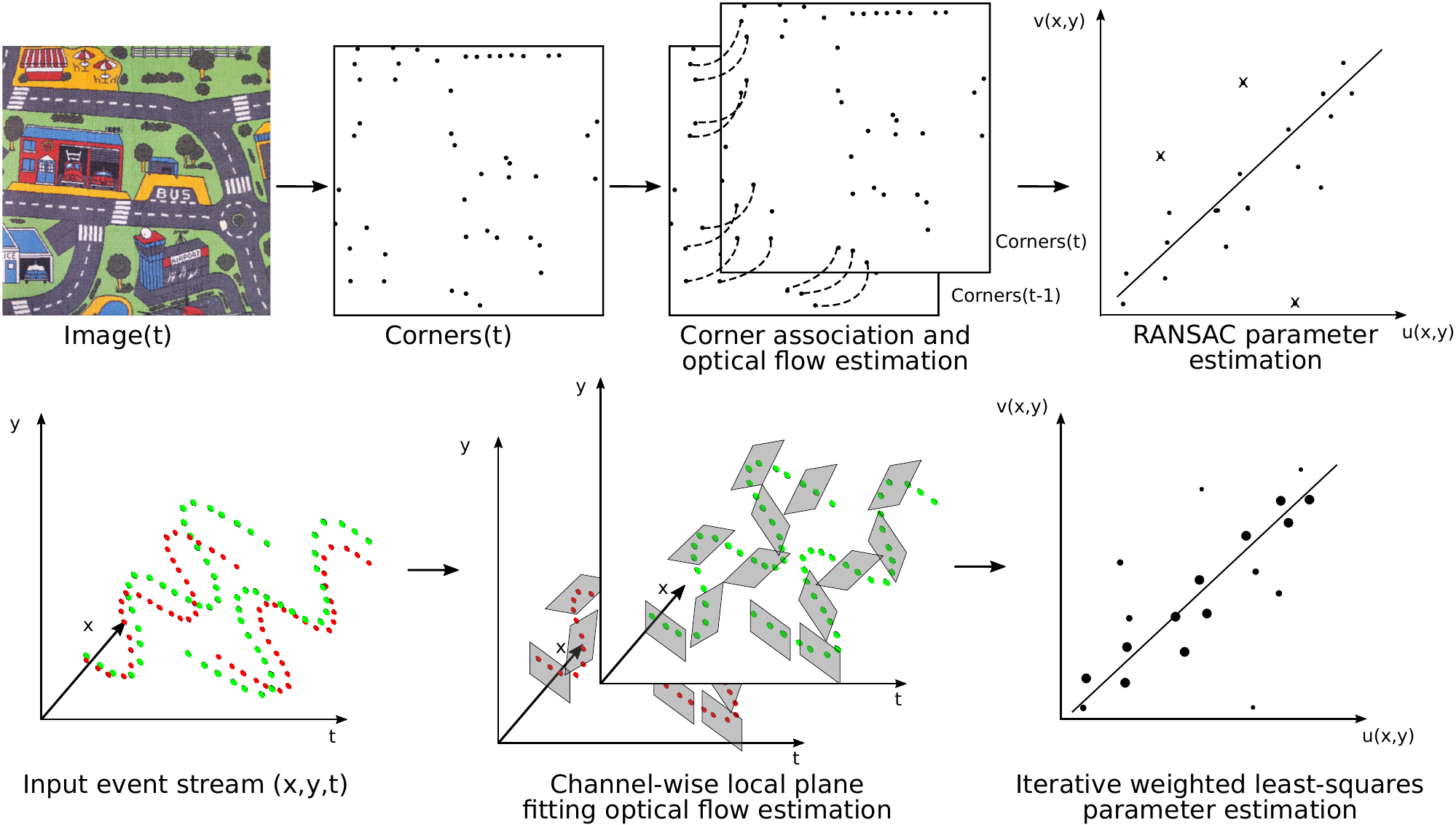}%
\caption{Process schematic of the optical flow computation for the CMOS camera (top) and the event-camera (below).}
\label{fig:process_flow}
\end{figure*}

The first camera uses the monochromatic information from the bottom looking CMOS camera built into the Parrot Bebop 2 using the size divergence estimation method described in \cite{Ho2018}. The second camera is the Insightness SEEM1 Dynamic Vision Sensor (DVS) using the efficient plane fitting optical flow estimation technique described in \cite{PijnackerHordijk2018}. This event based camera does not generate frames as a conventional image sensor but rather measures logarithmic light changes at each pixel independently and asynchronously. This makes the sensor conditioned to operate with high speed motion with low latency and relatively low data throughput. A simple comparison of the difference in data from the frame-based and event-based data can be seen in \cref{fig:frameVSevent}. This type of camera has been previously used to facilitate high speed landings as shown in \cite{PijnackerHordijk2018}. A schematic showing the optical flow processing pipeline for the CMOS and event-camera can be found in \cref{fig:process_flow}.

\begin{figure}[pos=!tb]
\centering
\includegraphics[width=0.45\textwidth]{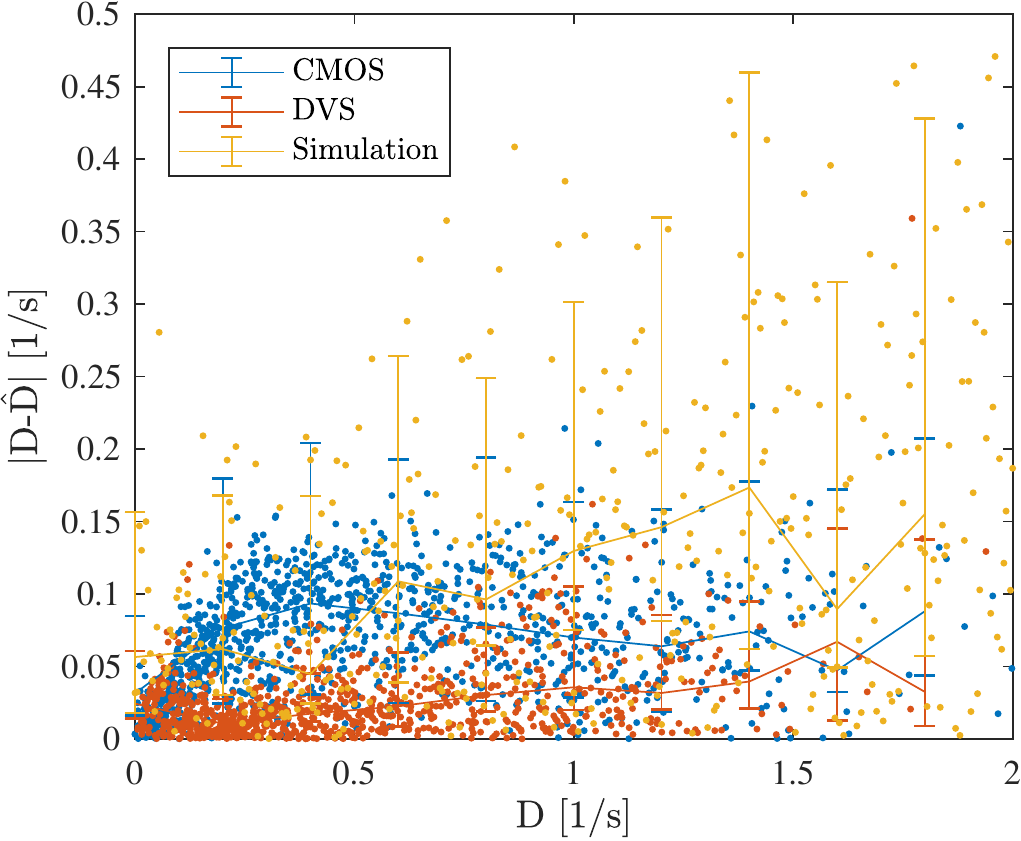}%
\caption{Divergence estimation error for the CMOS (blue) and DVS (red) cameras. The trend and distribution of these two cameras are quite different.}
\label{fig:reality_gap_div}
\end{figure}

\cref{fig:reality_gap_div} shows a comparison of the divergence estimation error from these two cameras highlighting how different the output of these camera is and how different they both are to the camera statistics used in simulation. Some additional differences in the camera properties are summarized in \cref{tab:camera_properties}. We will investigate in the next chapter if these differences result in a significant reality gap.

\begin{table}[pos=t]
    \centering
    \caption{Camera properties}
    \begin{tabular}{r|ccc}
        Property & Simulated & CMOS & DVS \\
        \hline
        Sensor & - & mt9v117 & SEES1 \\
        Resolution used & - & 240 x 240 & 262 x 262 \\
        Field of View & - & 58$^{\circ}$ & 79$^{\circ}$ \\
        Divergence Rate & [30, 50] Hz & 45 Hz & 100 Hz
    \end{tabular}
    \label{tab:camera_properties}
\end{table}
%
%
\section{Flight Test Results}
\label{sec:flightresults}

\begin{figure}[pos=!tb]
\centering
\includegraphics[width=0.45\textwidth]{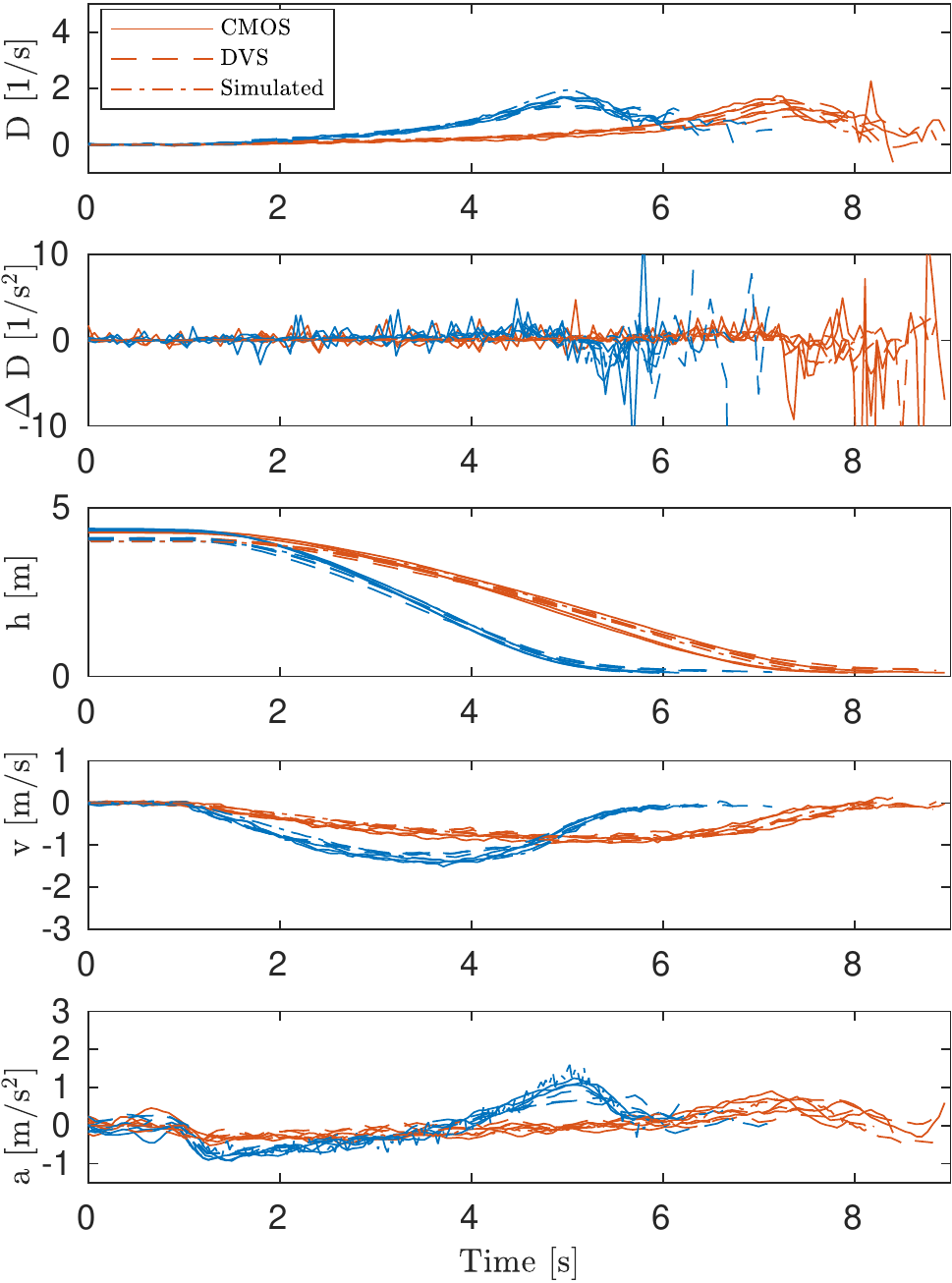}
\caption{Vehicle states and observations from real world flights with the NN$_1$ (blue) and NN$_2$ (red) controllers. The results using the CMOS camera is shown in solid and the DVS in dashed. The simulated performance is also plotted in dot-dash for comparison.}
\label{fig:real_nn}
\end{figure}

\begin{figure}[pos=!tb]
\centering
\includegraphics[width=0.45\textwidth]{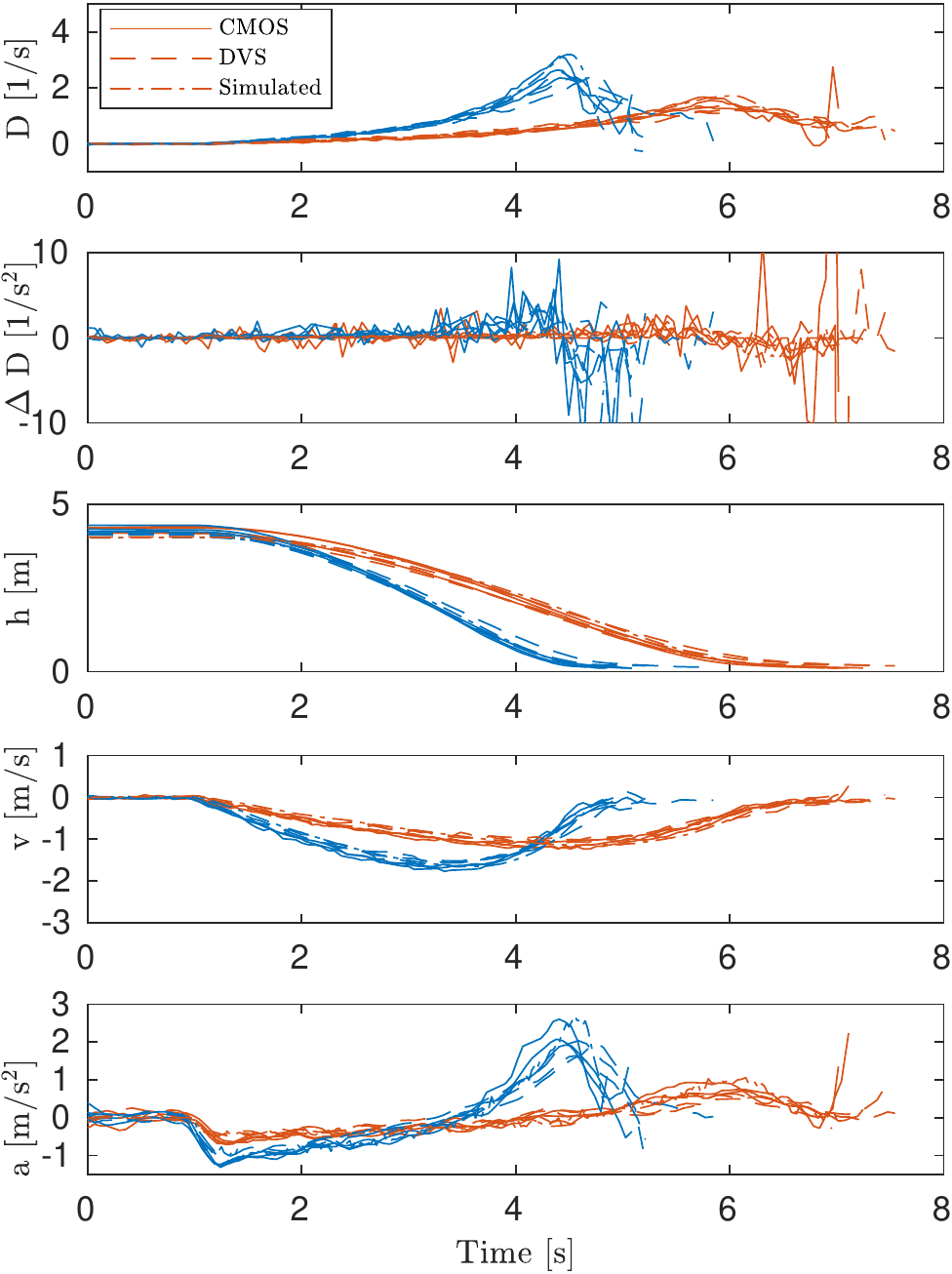}
\caption{Vehicle states and observations from real world flights with the RNN$_1$ (blue) and RNN$_2$ (red). The results using the CMOS camera is shown in solid and the DVS in dashed. The simulated performance is also plotted in dot-dash for comparison.}
\label{fig:real_rnn}
\end{figure}

\begin{figure}[pos=!tb]
\centering
\includegraphics[width=0.45\textwidth]{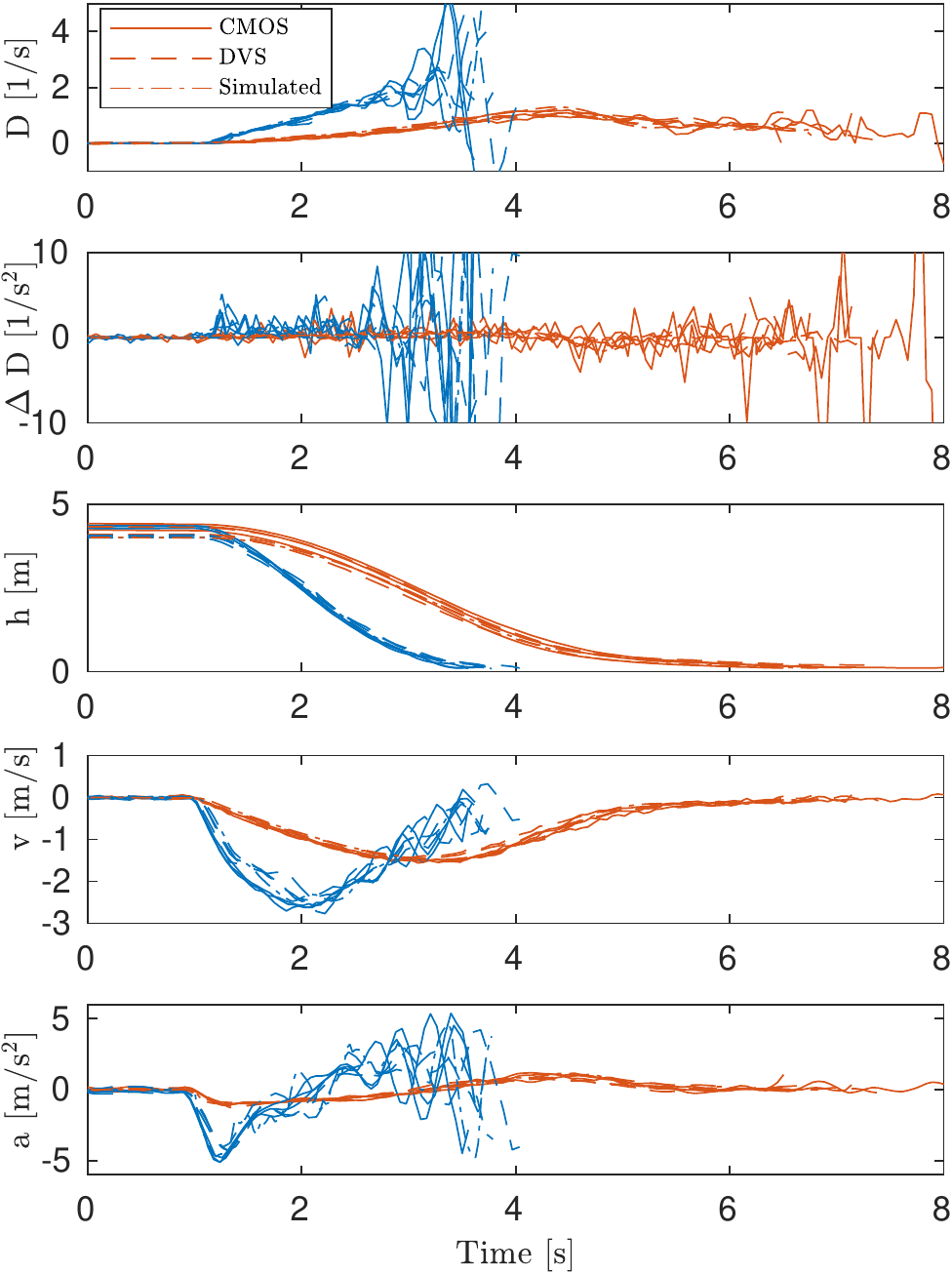}
\caption{Vehicle states and observations from real world flights with the CTRNN$_1$ (blue) and CTRNN$_2$ (red). The results using the CMOS camera is shown in solid and the DVS in dashed. The simulated performance is also plotted in dot-dash for comparison.}
\label{fig:real_ctrnn}
\end{figure}

Using the closed loop PI controller to control the thrust as described in the previous section, we performed a set of landings with some of the neurocontrollers identified in \cref{sec:evo}. All flights were initiated from a steady hover at an altitude of 4 m. \cref{fig:real_nn} shows the results from the controllers NN$_1$ and NN$_2$, \cref{fig:real_rnn} shows the results from RNN$_1$ and RNN$_2$ and \cref{fig:real_ctrnn} shows the results from CTRNN$_1$ and CTRNN$_2$. These results are plotted for both the CMOS and DVS cameras to see how well these two sensors affect the landing profile. The results from simulation have also been plotted to see how well the real world performance fits with the simulated, a measure for the eventual reality gap. NN$_3$, RNN$_3$ and CTRNN$_3$ were not tested in reality as their relatively high touchdown velocity in simulation may cause damage to our real world vehicle. This is in-fact a benefit of the multi-objective optimization scheme used here, the user can simply choose a policy that performs the trade-off of the fitness functions as desired. 

In spite of the differences in the generation of the divergence between the two cameras, the landing performance is very similar. These two systems are also so similar to the simulated landing that the plot is hardly visible. This would suggest that the eventual reality gap is small despite significant differences in the way the input was generated. 

Also notable is the repeatability of the landing maneuvers. The landings were performed three times each and each landing resulted in very similar trajectories. This shows that the evolutionary optimization converged to a robust solution as suggested by the analysis in \cref{sec:evo}.

%
%
\section{Conclusion}
\label{sec:conclusion}
This paper investigated the influence of abstraction of the sensory input to the reality gap for automatically optimized UAV agents tasked with performing quick yet safe landing. We have shown over multiple evolutionary runs and neurocontroller architectures, that abstraction does not unduly hamper the optimization power of the optimization as the agents developed a robust and effective method to land.

The optimized agents showed some landing strategies that were before not imagined by the human designers. One notable strategy is that instead of a simple proportional controller for the entire state space, an asymmetric response may be more appropriate to delay the onset of oscillations when performing the landing procedure. A strong response when the divergence error is positive and a weaker response when negative seems a good approach.

Tests in the real world showed the presence of significant differences between simulation and reality. The most significant was that the acceleration command tracking performance was poor, likely due to the drag and other non-linear aerodynamic effects which were not considered in simulation due to their modeling complexity. The resultant reality gap was crossed with the use of a closed loop controller representing another layer of abstraction, from the low-level raw motor control values to a desired vertical acceleration, which ensures robustness to the real world uncertainty.

Finally, we showed that abstraction on the sensory input of the neurocontroller was robust to the reality gap when using two different input estimation techniques. Although two cameras with different imaging techniques were used, the resultant landing profile was very similar. Abstraction can therefore be a powerful tool when crossing the reality gap.

Future work could investigate how this approach can generalize to different vehicles operating in different real world environments. This would help to demonstrate the effectiveness of the approach to not only cross the reality gap but to address the more general transfer problem.

\section*{Acknowledgment}
We would like to thank the team at Insightness AG for their assistance in porting the SEEM1 SDK to operate on the ARM processor of the Parrot Bebop 2. Without their assistance the results in the paper would not be possible.

\vfill\null

\section*{Appendix: Controller Steady State Input-Output Mappings}

\vfill\null

\bibliography{library}

\bio[pos=l]{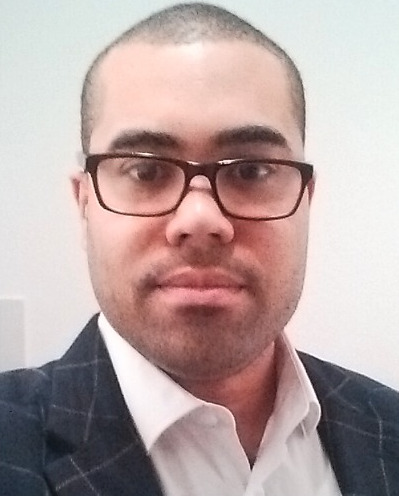}
\textbf{Kirk~Y.~W.~Scheper} received his M.Sc. and Ph.D from the Micro Air Vehicle Laboratory in the Faculty of Aerospace Engineering at Delft University of Technology, the Netherlands, in 2014 and 2019 respectively. His research focuses on the development of embedded software which facilitates high level autonomy of micro air vehicles. His work is mainly in the fields of evolutionary robotics, embodied cognition, and vision-based navigation and event-based cameras.
\endbio

\bio[pos=l]{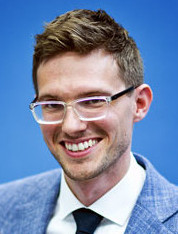}
\textbf{Guido~C.~H.~E.~de~Croon} received his M.Sc. and Ph.D. in the field of Artificial Intelligence at Maastricht University, the Netherlands. His research interest lies with computationally efficient algorithms for robot autonomy, with an emphasis on computer vision. Since 2008 he has worked on algorithms for achieving autonomous flight with small and light-weight flying robots, such as the DelFly flapping wing MAV. In 2011-2012, he was a research fellow in the Advanced Concepts Team of the European Space Agency, where he studied topics such as optical flow based control algorithms for extraterrestrial landing scenarios. Currently, he is associate professor at Delft University of Technology, the Netherlands, where he is the scientific lead of the Micro Air Vehicle Laboratory.
\endbio


\vfill


\end{document}